\documentclass[onefignum,onetabnum]{siamart171218}

\usepackage{lipsum}
\usepackage{amsfonts}
\usepackage{graphicx}
\usepackage{epstopdf}

\usepackage{multirow}
\usepackage{ulem}
\usepackage{tikz}
\usetikzlibrary{matrix}
\usetikzlibrary{shapes,patterns,snakes,arrows}
\usetikzlibrary{positioning,fit,calc}
\usetikzlibrary{graphs}

\definecolor{dblue}{RGB}{0,0,139}
\definecolor{dred}{rgb}{0.8,0,0}

\usepackage{algorithmic}
\ifpdf
  \DeclareGraphicsExtensions{.eps,.pdf,.png,.jpg}
\else
  \DeclareGraphicsExtensions{.eps}
\fi

\newsiamremark{remark}{Remark}
\newsiamremark{hypothesis}{Hypothesis}
\crefname{hypothesis}{Hypothesis}{Hypotheses}
\newsiamthm{claim}{Claim}

\def\bb{\begin{bmatrix}}
\def\eb{\end{bmatrix}}

\newcommand{\TheTitle}{A Domain Decomposition-Based CNN-DNN Architecture for Model Parallel Training Applied to Image Recognition Problems}

\headers{A Domain Decomposition-Based CNN-DNN Architecture}{A. Klawonn, M. Lanser, and J. Weber}

\title{{\TheTitle}}

\author{Axel Klawonn\footnotemark[2]\ \footnotemark[3] \and Martin Lanser\footnotemark[2]\ \footnotemark[3] \and Janine Weber\footnotemark[2]\ \footnotemark[3]}

\usepackage{amsopn}

\usepackage{ifthen}%
\newcommand{\mycomment}[1]{%
\ifthenelse{\isodd{\value{page}}}{%
\normalmarginpar%
\marginpar{\tiny {#1}}%
}{%
\reversemarginpar%
\marginpar{\tiny {#1}}%
}}%

\usepackage{cite}
\usepackage[english]{babel}
\usepackage{hyperref} 
\usepackage{amsmath}
\usepackage{amssymb}
\usepackage{amsfonts}
\usepackage[T1]{fontenc}
\usepackage[utf8]{inputenc}

\usepackage{makeidx}  % zur Einbindung von Grafiken und Abbildungen
\usepackage{enumerate}
\usepackage{pgfplots}
\usepackage{tikz}
\usetikzlibrary{matrix}
\usetikzlibrary{shapes,patterns,snakes,arrows,decorations.markings,quotes,arrows.meta}
\usetikzlibrary{positioning,fit,calc}
\usetikzlibrary{3d} %for including external image 
\usetikzlibrary{graphs}
\usepackage{ifthen}
\usepackage{nicefrac}
\usepackage{relsize}
\usepackage{bbm}  % um die Indikatorfkt zu schreiben

\usepackage{listings}  % um Wuellcode korrekt darstellen zu koennen
\usepackage{booktabs} % um Tabellen schoener darzustellen

\definecolor{darkspringgreen}{rgb}{0.09, 0.45, 0.27}
\definecolor{bostonuniversityred}{rgb}{0.8, 0.0, 0.0}
\definecolor{ferrarired}{rgb}{1.0, 0.11, 0.0}
\definecolor{coralred}{rgb}{1.0, 0.25, 0.25}
\definecolor{pastelred}{rgb}{1.0, 0.41, 0.38}
\definecolor{darkblue}{RGB}{0,0,139}
\definecolor{darkred}{RGB}{105,0,0}

\def\ConvColor{rgb:yellow,5;red,2.5;white,5}
\def\ConvReluColor{rgb:yellow,5;red,5;white,5}
\def\PoolColor{rgb:red,1;black,0.3}
\def\FcColor{rgb:blue,2;green,1;black,0.1}
\def\FcReluColor{rgb:blue,2;green,1;black,0.2}
\def\SoftmaxColor{rgb:cyan,5;black,3}

\def\edgecolor{rgb:blue,4;red,1;green,4;black,3}
\newcommand{\midarrow}{\tikz \draw[-Stealth,line width =0.8mm,draw=\edgecolor] (-0.3,0) -- ++(0.3,0);}

\usepackage{Ball}
\usepackage{Box}
\usepackage{RightBandedBox}

\def\x{{\bf x}}

\newcommand{\VGDSW}[1]{V_{\rm{GDSW}}}

\newcommand{\takeout}[1]{ }  % Nix, nothing

\tikzset{%
  every neuron/.style={
    circle,
    draw,
    minimum size=0.6cm
  },
  every input neuron/.style={
    circle,
    draw,
    minimum size=0.6cm,
    fill={rgb:cyan,3;black,1}
  },
  every output neuron/.style={
    circle,
    draw,
    minimum size=0.6cm,
    fill=orange!30
  },
  every hidden neuron/.style={
    circle,
    draw,
    minimum size=0.6cm,
    fill={rgb:blue,1;green,0.5;black,0.1}
  },
  neuron missing/.style={
    draw=none, 
    scale=1.5,
    text height=0.3cm,
    execute at begin node=\color{black}$\vdots$
  },
}

\begin{document}

\maketitle

\renewcommand{\thefootnote}{\fnsymbol{footnote}}

\footnotetext[2]{Department of Mathematics and Computer Science, University of Cologne, Weyertal 86-90, 50931 K\"oln, Germany, \email{\{axel.klawonn, martin.lanser, janine.weber\}@uni-koeln.de}, url: \url{http://www.numerik.uni-koeln.de}} 
\footnotetext[3]{Center for Data and Simulation Science, University of Cologne, Germany, url: \url{http://www.cds.uni-koeln.de}}

% REQUIRED
\begin{abstract}
Deep neural networks (DNNs) and, in particular, convolutional neural networks (CNNs) have brought significant advances in a wide range of modern computer application problems. 
However, the increasing availability of large amounts of datasets as well as the increasing available computational power of modern computers lead to a steady growth in the complexity and size of DNN and CNN models, respectively, and thus, to longer training times. 
Hence, various methods and attempts have been developed to accelerate and parallelize the training of complex network architectures.
In this work, a novel CNN-DNN architecture is proposed that naturally supports a model parallel training strategy and that is loosely inspired by two-level domain decomposition methods (DDM). 
First, local CNN models, that is, subnetworks, are defined that operate on overlapping or nonoverlapping parts of the input data, for example, sub-images. 
The subnetworks can be trained completely in parallel and independently of each other. Each subnetwork then outputs a local decision for the given machine learning problem which is exclusively based on the respective local input data. 
Subsequently, in a second step, an additional DNN model is trained which evaluates the local decisions of the local subnetworks and generates a final, global decision. With respect to the analogy to DDM, the DNN can be loosely interpreted as a coarse problem and hence, the new approach can be interpreted as a two-level domain decomposition. 
In this paper, we apply the proposed architecture to image classification problems using CNNs.  
 Experimental results for different 2D image classification problems are provided as well as a face recognition problem, and a classification problem for 3D computer tomography (CT) scans. Therefore, classical ResNet and VGG architectures are considered. More modern architectures, as, e.g., MobileNet2, are left for future work. 
The results show that the proposed approach can significantly accelerate the required training time compared to the global model and, additionally, can also help to improve the accuracy of the underlying classification problem. 
\end{abstract}

% REQUIRED

% REQUIRED

\section{Introduction}
\label{sec:intro}

Supervised machine learning and, in particular, deep neural networks (DNNs)~\cite{lecun2015deep,Goodfellow:2016:DL} have become an important tool in various modern computer applications and have brought significant advances for a wide range of problems, such as, for example, image classification~\cite{simonyan:2014:VGGnet,he2016deep}, object detection~\cite{lecun1989backpropagation}, face recognition~\cite{parkhi:2015:deepface}, medical image diagnosis~\cite{singh:2020:3d_med_review}, and many other applications.
Among the many different types of neural networks, convolutional neural networks (CNNs)~\cite{lecun:1989:CNN} are tremendously successful in processing data with a grid-like topology such as, for example, time series data, or 2D and 3D grids of pixels and voxels, respectively~\cite{Goodfellow:2016:DL,chollet2017deep}. 
With the increasing available computational power and the rising popularity of CNNs and DNNs, in general, the trained network models tend to steadily grow in their size and complexity. In combination with the increasing availability of large amounts of datasets, this leads to increasingly longer training times and higher memory demands such that high-performing computing clusters are usually required for the development of these models.
Hence, various methods and attempts are developed to accelerate and parallelize the training of complex network architectures.

In general, parallelism and distribution in the context of neural networks can be roughly categorized into model parallelism, data parallelism, pipelining, and hybrid approaches~\cite{ben2019demystifying}.
In the following, we briefly describe the main idea of the different approaches. A more detailed overview can be found in, for example,~\cite{ben2019demystifying} and the references therein. 
In data parallelism, the training dataset is partitioned into different subsets and each parallel processor or different parallel device has access to exactly one of these subsets.  
Additionally, each processor has a local copy of the entire network model which is exclusively trained with the local training data.
A detailed investigation of the effects of data parallelism on the training of neural networks is, e.g., given in~\cite{shallue2018measuring}.
In model parallelism, the neurons in each layer are partitioned into subsets of the network which are trained in parallel on different processors or devices~\cite{forrest1987implementing}. This can conserve the memory requirements but usually requires communication between the different processors due to the dependency of the neurons of a given layer on all neurons of the previous layer. 
In pipelining, a DNN model is usually partitioned according to its depth such that the layers of the network are assigned to different processors~\cite{ben2019demystifying}. 
Hence, pipelining can be interpreted as a form of model parallelism since the length of the pipeline depends on the given DNN structure. Additionally, due to the distribution of the DNN layers to different processors, samples are processed through the network in parallel such that pipelining can also be interpreted as a form of data parallelism~\cite{ben2019demystifying}.
Moreover, a wide range of hybrid approaches exists in order to combine the advantages of the mentioned classes of parallelism by compensating each of their drawbacks. 

In the recently growing field of scientific machine learning~\cite{USDpt:SciML}, existing methods from supervised or unsupervised machine learning are combined with expertise from different research areas such as, for example, numerical simulations or iterative solvers in order to develop new, hybrid methods which benefit from the knowledge in both areas. 
In this work, we propose a novel network architecture for image recognition problems which is inspired by our experience with domain decomposition methods and which naturally supports a model parallel training strategy.
Domain decomposition methods~\cite{toselli,QuarteroniValli2008} are iterative, highly scalable methods for the solution of partial differential equations. They rely on a divide-and-conquer principle, that is, they decompose the global problem into a number of smaller, local subproblems which are solved in parallel. In order to obtain a global, continuous solution and to ensure a fast transport of information between the different subproblems, in so-called two-level domain decomposition methods, an additional global \textit{coarse problem} is solved in each iteration. 
In~\cite{GuCai:2022:dd_transfer}, the idea of domain decomposition methods has been transferred to the training of CNNs.  There, the authors propose to decompose a global CNN along its width into a finite number of smaller, local subnetworks which can be trained independently. The obtained weights of the local subnetworks are then used as an initialization for the subsequently trained global network by using a transfer learning strategy. In~\cite{li2021summation}, the authors propose a local principal component analysis (PCA) which also relies on the ideas of domain decomposition methods.
In~\cite{dolean2022finite,shukla2021parallel,yang2022additive,wu2022improved}, various decomposed training strategies that are inspired by different domain decomposition methods have been introduced for physics-informed neural networks~\cite{RPK:2019:MID} by several authors. 
 In~\cite{kopanicakova2022globally}, a multilevel training strategy for deep residual neural networks is proposed. 
A detailed overview of other existing work on combining domain decomposition or multilevel methods with machine learning can be found in the review paper~\cite{HKLW:2020:GammReview}.

In the present work, we propose a new CNN-DNN architecture which is loosely inspired by two-level domain decomposition methods and that can be implemented employing a model parallel training strategy. Let us remark that in contrast to a classical model parallel training strategy, the given architecture of the considered network is actually changed to some degree in order to improve the parallel scalability and performance of the model. First, similarly to~\cite{GuCai:2022:dd_transfer}, a given global CNN is decomposed into a number of smaller, local CNNs, that is, subnetworks which can be trained completely in parallel and independently of each other. Each of these subnetworks operates on all samples of the dataset but, correspondingly to the decomposition of the CNN, only on a specific part of each sample. In particular, each local CNN is trained with a separate loss function which measures the difference between the local classifications based on the sub-data and the underlying global class labels. As a consequence, the strength of our approach is that it splits large individual data points as high-resolution images into smaller ones and thus, in contrast to data parallel approaches, is especially well suited for high resolution data or 3D data. In the present paper, we restrict ourselves to the training of CNNs for 2D and 3D image classification problems. We also restrict ourselves to classical CNNs as, for example, ResNets (Residual Networks)~\cite{he2016deep} and VGG~\cite{simonyan:2014:VGGnet}. For a more detailed study of the capability and parallel performance of our model certainly more modern architectures, as, e.g., current versions of MobileNet~\cite{howard2017mobilenets,sandler2018mobilenetv2}, should be considered in the future. 
 Within our proposed approach, all images of a given dataset are decomposed into overlapping or nonoverlapping sub-images and each subnetwork operates on a set of these sub-images. Then, in a second step, we train an additional DNN model which evaluates the local classifications of the different subnetworks into a final, global decision or classification, respectively. With respect to the analogy to domain decomposition methods, this DNN can be loosely interpreted as a \textit{global coarse problem} and we refer to our approach as a two-level domain decomposition.
  To clarify the scope of this paper, we would like to highlight that, within this paper, it is not our primary goal to present a novel deep learning model that provides the highest possible accuracy values for different challenging image classification problems. Instead, our intention is to provide a model parallel training approach for a given CNN model in order to accelerate the training process on a parallel GPU cluster. Additionally, we compare the accuracy and generalization properties for a given global CNN model and the corresponding model parallel approach in order to ensure that we fully benefit from the accelerated training without loosing too much classification accuracy.
We provide experimental results for two different 2D image recognition problems~\cite{Cifar10_TR,tfflowers}, a face recognition problem~\cite{Yale_faces1,Yale_faces2} and a classification problem based on 3D chest computer tomography (CT) scans~\cite{Chest_CT}. 
Let us note that, in general, the presented approach is not restricted to CNNs for image recognition problems and could also be extended to other network architectures than CNNs and different image-based application problems. 

The remainder of the paper is organized as follows. 
First, in~\cref{sec:classic_cnn}, we define our problem setting which is the classification of 2D or 3D image data using CNNs~\cite{lecun:1989:CNN}. Additionally, we provide a short mathematical introduction into CNNs and the related mathematical operations.
We present the main algorithmic part of the present paper in~\cref{sec:algo} such that it includes the definition of our proposed hybrid CNN-DNN architecture and the corresponding model parallel training strategy.
 Next, we provide a detailed description of our test datasets in~\cref{sec:data_nets}. We test our proposed network model and the related training strategy for three different two-dimensional image datasets, that is, pixel data, and  one three-dimensional image dataset corresponding to CT data, that is, voxel data.
In~\cref{sec:res}, we provide experimental results for the proposed CNN-DNN architecture and the different image classification or recognition problems. We provide classification accuracies of our approach in direct comparison to classic, global CNN models as well as runtimes required for the training of the different models.
In particular, for the 3D chest CT CNN model, we are able to reduce the required training time up to a factor of almost $23$ using our model parallel strategy compared to the respective globally trained 3D CNN model. 
All experiments have been performed on a workstation with $8$ NVIDIA Tesla V100 32GB GPUs using python 3.6 and TensorFlow 2.4~\cite{tensorflow2015-whitepaper}.
Finally, we draw a conclusion and present possible future research directions in~\cref{sec:concl}.

\section{Problem Setting: Image Classification with Convolutional Neural Networks}
\label{sec:classic_cnn}

In this paper, we apply our proposed parallel training approach to different image classification problems, for example, synthetic image recognition tasks, face recognition, and medical imaging.
Given that convolutional neural networks (CNNs)~\cite{lecun:1989:CNN} are tremendously successful in processing 2D or 3D image data, see, for example,~\cite[Chapt. 9]{Goodfellow:2016:DL}, we train different types of 2D or 3D CNN models for the different image classification problems.
In this section, we briefly describe the basic structure of a classical CNN model. The following descriptions are roughly based on~\cite[Chapt. 9]{Goodfellow:2016:DL} and~\cite[Chapt. 5]{chollet2017deep}.

CNNs are a special type of neural networks and, to some extent, generalize dense feedforward neural networks by using convolutional operators instead of general matrix multiplications in at least one of their layers~\cite[Chapt. 9]{Goodfellow:2016:DL}. Due to their specific structure, CNNs are specialized for processing data with a grid-like topology such as, for example, time series data, or 2D and 3D grids of pixels and voxels, respectively. 
In general, the structure of a CNN can be described by two basic operations or types of layers, that is, the already mentioned convolutional layers as well as pooling layers. Both are a specific kind of linear operations.

\begin{figure}[th]
\centering
\includegraphics[width=0.7\textwidth]{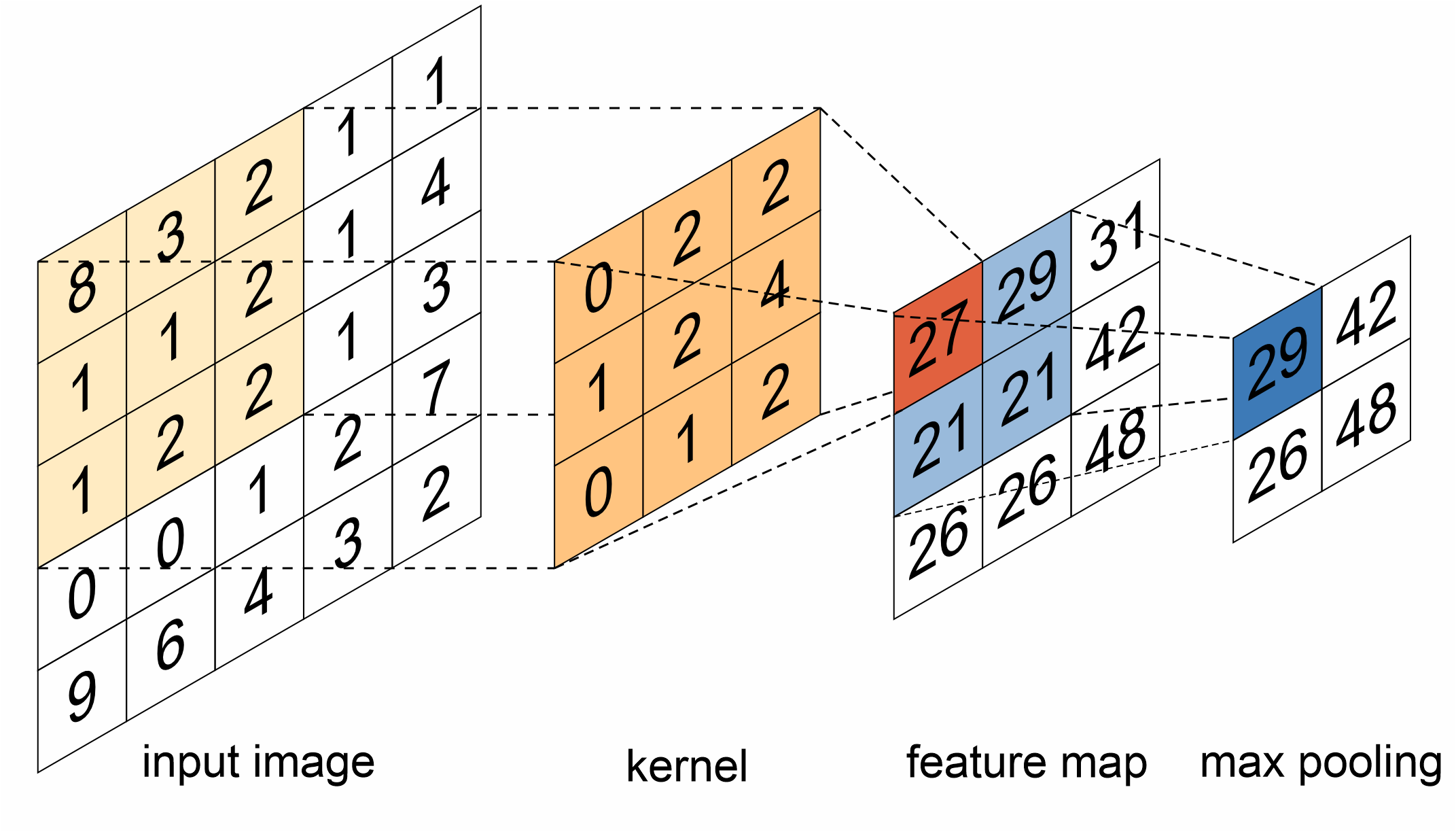}
\caption{Visualization of a convolution (left, in orange) and a maximum pooling (right, in blue) operation within a CNN. For the convolution, we use a kernel of size $3\times3$ pixels without padding and a stride equal to one. For the pooling, we use a window size of $2\times 2$ pixels.}
\label{fig:conv_pool}
\end{figure}

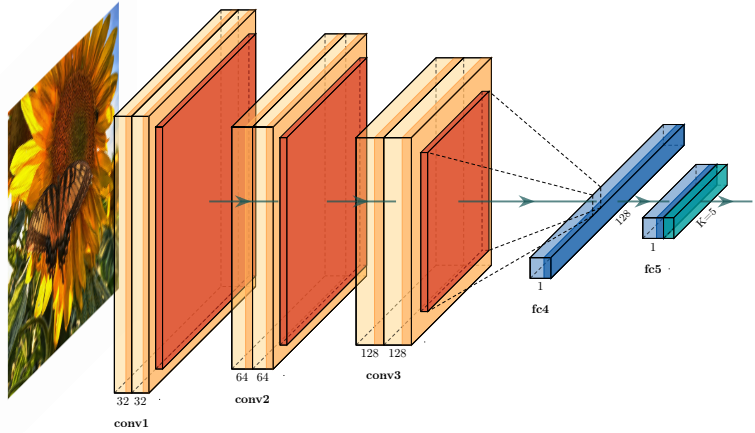
\begin{figure}[t]
\centering
\scalebox{0.46}{

\begin{tikzpicture}
\tikzstyle{connection}=[ultra thick,every node/.style={sloped,allow upside down},draw=\edgecolor,opacity=0.7]
%%%%%%%%%%%%%%%%%%%%%%%%%%%%%%%%%%%%%%%%%%%%%%%%%%%%%%%%%%%%%%%%%%%%%%%%%%%%%%%%%%%%%%%%
%% Draw Layer Blocks
%%%%%%%%%%%%%%%%%%%%%%%%%%%%%%%%%%%%%%%%%%%%%%%%%%%%%%%%%%%%%%%%%%%%%%%%%%%%%%%%%%%%%%%%
\node[canvas is zy plane at x=0] (temp) at (-3,0,0) {\includegraphics[width=11cm,height=11cm]{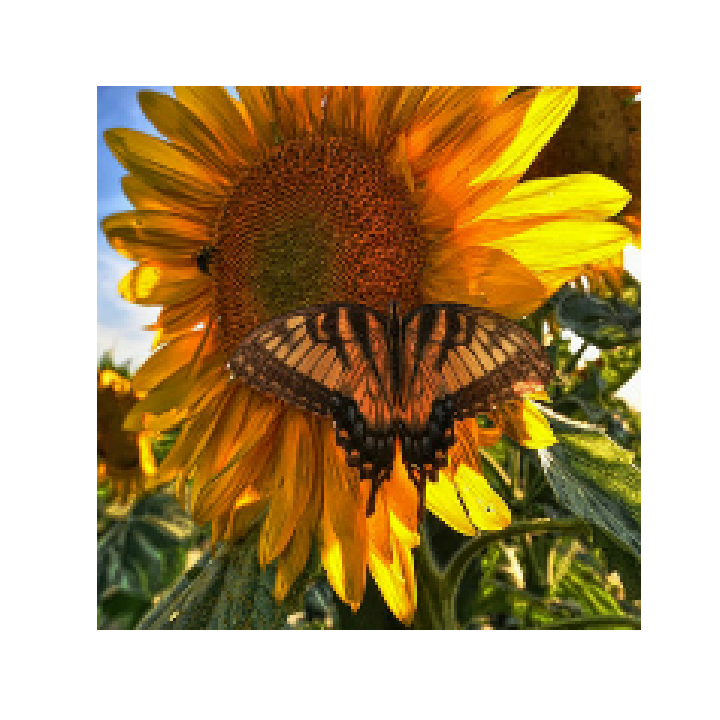}};
% conv1_1,conv1_2
\pic[shift={(0,0,0)}] at (0,0,0) {RightBandedBox={name=cr1,caption=conv1,%
        xlabel={{"32","32"}},fill=\ConvColor,bandfill=\ConvReluColor,%
        height=40,width={2.5,2.5},depth=40}};
%pool1
\pic[shift={(0,0,0)}] at (cr1-east) {Box={name=p1,caption=,%
        fill=\PoolColor,opacity=0.5,height=35,width=1,depth=35}};
%%%%%%%%%%
% conv2_1,conv2_2
\pic[shift={(2,0,0)}] at (p1-east) {RightBandedBox={name=cr2,caption=conv2,%
        xlabel={{"64","64"}},fill=\ConvColor,bandfill=\ConvReluColor,%
        height=35,width={3,3},depth=35}};
%pool2
\pic[shift={(0,0,0)}] at (cr2-east) {Box={name=p2,caption=,%
        fill=\PoolColor,opacity=0.5,height=30,width=1,depth=30}};
%%%%%%%%%%
% conv3_1,conv3_2
\pic[shift={(2,0,0)}] at (p2-east) {RightBandedBox={name=cr3,caption=conv3,%
        xlabel={{"128","128"}},fill=\ConvColor,bandfill=\ConvReluColor,%
        height=30,width={4,4},depth=30}};
%pool3
\pic[shift={(0,0,0)}] at (cr3-east) {Box={name=p3,caption=,%
        fill=\PoolColor,opacity=0.5,height=23,width=1,depth=23}};
%%%%%%%%%%
% conv4_1,conv4_2,conv4_3
%\pic[shift={(1.8,0,0)}] at (p3-east) {RightBandedBox={name=cr4,caption=conv4,%
%        xlabel={{"512","512","512"}},zlabel=28,fill=\ConvColor,bandfill=\ConvReluColor,%
%        height=23,width={7,7,7},depth=23}};
%%pool4
%\pic[shift={(0,0,0)}] at (cr4-east) {Box={name=p4,caption=,%
%        fill=\PoolColor,opacity=0.5,height=15,width=1,depth=15}};
%%%%%%%%%%
% conv5_1,conv5_2,conv5_3
%\pic[shift={(1.5,0,0)}] at (p4-east) {RightBandedBox={name=cr5,caption=conv5,%
%        xlabel={{"512","512","512"}},zlabel=14,fill=\ConvColor,bandfill=\ConvReluColor,%
%        height=15,width={7,7,7},depth=15}};
%%pool5
%\pic[shift={(0,0,0)}] at (cr5-east) {Box={name=p5,caption=,%
%        fill=\PoolColor,opacity=0.5,height=10,width=1,depth=10}};
%%%%%%%%%%
% fc6
\pic[shift={(4,0,0)}] at (p3-east) {RightBandedBox={name=fc4,caption=fc4,%
        xlabel={{"1",""}},zlabel=128,fill=\FcColor,bandfill=\FcReluColor,%
        height=3,width=3,depth=50}};
%%%%%%%%%%
% fc7
%\pic[shift={(2,0,0)}] at (fc6-east) {RightBandedBox={name=fc7,caption=fc7,%
%        xlabel={{"1","dummy"}},zlabel=4096,fill=\FcColor,bandfill=\FcReluColor,%
%        height=3,width=3,depth=100}};
%%%%%%%%%%
% fc8
\pic[shift={(1.5,0,0)}] at (fc4-east) {RightBandedBox={name=fc5,caption=fc5,%
        xlabel={{"1","dummy"}},fill=\FcColor,bandfill=\FcReluColor,%
        height=3,width=3,depth=20}};

%%%%%%%%%%
% softmax
\pic[shift={(0,0,0)}] at (fc5-east) {Box={name=softmax,caption=,%
        xlabel={{"","dummy"}},zlabel={K=5},opacity=0.8,fill=\SoftmaxColor,%
        height=3,width=1.5,depth=20}};
    
%%%%%%%%%%%%%%%%%%%%%%%%%%%%%%%%%%%%%%%%%%%%%%%%%%%%%%%%%%%%%%%%%%%%%%%%%%%%%%%%%%%%%%%%
%% Draw Arrow Connections
%%%%%%%%%%%%%%%%%%%%%%%%%%%%%%%%%%%%%%%%%%%%%%%%%%%%%%%%%%%%%%%%%%%%%%%%%%%%%%%%%%%%%%%%
\draw [connection]  (p1-east)        -- node {\midarrow} (cr2-west);
\draw [connection]  (p2-east)        -- node {\midarrow} (cr3-west);
%\draw [connection]  (p3-east)        -- node {\midarrow} (cr4-west);
%\draw [connection]  (p4-east)        -- node {\midarrow} (cr5-west);
\draw [connection]  (p3-east)        -- node {\midarrow} (fc4-west);
\draw [connection]  (fc4-east)       -- node {\midarrow} (fc5-west);
%\draw [connection]  (fc7-east)       -- node {\midarrow} (fc8-west);
\draw [connection]  (softmax-east)   -- node {\midarrow} ++(1.5,0,0);
%%%%%%%%%%%%%%%%%%%%%%%%%%%%%%%%%%%%%%%%%%%%%%%%%%%%%%%%%%%%%%%%%%%%%%%%%%%%%%%%%%%%%%%%
%% Draw Dotted Edges 
%%%%%%%%%%%%%%%%%%%%%%%%%%%%%%%%%%%%%%%%%%%%%%%%%%%%%%%%%%%%%%%%%%%%%%%%%%%%%%%%%%%%%%%%
\draw[densely dashed]
    (fc4-west)++(0, 1.5*.2, 1.5*.2) coordinate(a) -- (p3-nearnortheast)
    (fc4-west)++(0,-1.5*.2, 1.5*.2) coordinate(b) -- (p3-nearsoutheast)
    (fc4-west)++(0,-1.5*.2,-1.5*.2) coordinate(c) -- (p3-farsoutheast)
    (fc4-west)++(0, 1.5*.2,-1.5*.2) coordinate(d) -- (p3-farnortheast)
    
    (a)--(b)--(c)--(d)
    ;
%%%%%%%%%%%%%%%%%%%%%%%%%%%%%%%%%%%%%%%%%%%%%%%%%%%%%%%%%%%%%%%%%%%%%%%%%%%%%%%%%%%%%%%%
\end{tikzpicture}

}
\caption{\textit{Global}, undecomposed CNN (VGG3 type~\cite{simonyan:2014:VGGnet}) used for the image classification of the TF-Flowers dataset~\cite{tfflowers}. 
We visualize the different convolutional (conv) and fully connected layers (fc) using the following color code. We mark in light orange a convolutional layer with a subsequent ReLU activation layer. We mark in dark orange a maximum pooling layer. In blue, we mark a fully connected layer with a subsequent ReLU activation layer and in turquoise, we mark a softmax activation layer. Visualization using~\cite{CNN_repo}.}
\label{fig:cnn}
\end{figure}

\paragraph{Convolutional layers}
In the context of deep learning, the term \textit{convolution} refers to a discrete convolution of two discrete functions. Mathematically, this can be written as a multiplication of a tensor, for example, a 2D or 3D tensor, by a \textit{kernel} matrix. Basically, this corresponds to a weighted average function that is applied to the input tensor where the discrete weights $w$ in form of the kernel are learned by the network during training. 
In CNNs, the kernel has usually a much smaller size than the input image. Thus, CNNs typically have sparse interactions and are very effective at extracting local, meaningful features from the input data; see~\cite[Chapt. 9]{Goodfellow:2016:DL} and~\cite[Chapt. 5]{chollet2017deep}. The output of a convolutional layer is then obtained by sliding the kernel over patches of the input tensor, also called \textit{feature map}, and successively computing the weighted sum of the kernel entries and the underlying values along these patches. The resulting output is often referred to as the \textit{output feature map}. In~\cref{fig:conv_pool} (left), we show an exemplary convolutional operation with a kernel size of $3\times 3$ pixels applied to a 2D pixel image. 
In general, one typically uses multiple kernels or \textit{filters} on the feature maps of a CNN which also operate on the different \textit{channels} of the input data. For example, we have three channels for an RGB image. 
Depending on the type of a convolutional layer which is characterized by the type of \textit{padding} it uses, its output width and height can differ from the input width and height. Generally speaking, when using neither padding nor \textit{striding} in a convolutional layer, the output feature map will shrink with each convolutional layer given that you cannot fully center the kernel around the border patches of the feature map; see also~\cref{fig:conv_pool}.
In order to obtain an output feature map with the same width and height as the input feature map, one can use padding. Padding basically adds an appropriate number of rows and columns on each side of the input feature map such that it is possible to fit the center kernel windows also around the border patches of the original input feature map. Striding, however, can have the opposite effect since it increases the shift of the local kernels within the convolutions such that the data dimension is reduced even further. In~\cref{fig:cnn}, we show the different types of layers of the chosen CNN architecture as boxes of different height and width indicating the different output dimensions and numbers of channels of the different convolutional layers. 
See also~\cite[Chapt. 9]{Goodfellow:2016:DL} and~\cite[Chapt. 5]{chollet2017deep} for more details on different types of padding and striding.
Let us note that, usually, a stack of convolutional layers is followed by a nonlinear activation function in order to learn a highly nonlinear functional relation between the input and the output data of the CNN.

\paragraph{Pooling layers}
Typically, a stack of convolutional layers and a subsequent nonlinear activation function is followed by a \textit{pooling function} or a pooling layer. A pooling layer replaces the output at a certain location of the previous layer by a summary statistic of the nearby outputs. Common examples for such pooling operations are the maximum pooling operation or the average pooling operation. The latter replaces a rectangular neighborhood with its average value whereas the former replaces it with its maximum value. In both cases, pooling helps to further downsample the respective feature map and, in particular, makes the feature map more invariant to small translations of the input~\cite[Chapt. 9]{Goodfellow:2016:DL}.
In~\cref{fig:conv_pool} (right), we show an exemplary maximum pooling operation with a window size of $2\times 2$ pixels.

To conclude this section, let us note that for the CNN models in~\cref{sec:data_nets}, we always complete the CNNs with one or two fully connected layers. In particular, the last layer 
has always as many neurons as we have classes in the underlying classification problem and we always use a sigmoid activation function for the last fully connected layer. Hence, the trained CNN models always map the input, that is, either 2D pixel or 3D voxel data to a corresponding probability distribution among the $K$ different classes of the discrete classification problem.

\section{A Domain Decomposition-based CNN-DNN Architecture for Model Parallel Training}
\label{sec:algo}

In this section, we describe the core idea of our proposed model parallel training approach for CNN models. Combining the ideas of CNNs and DNNs, we suggest a completely new class of network architectures which naturally supports parallel training and builds on the strength of well-established CNN architectures for different applications.
Let us note again that, in general, CNNs are very effective in processing data with a grid-like structure such as, e.g., time series data or pixel or voxel data. However, with the increasing availability of large amounts of data in various applications such as, for example, face recognition or medical imaging~\cite{parkhi:2015:deepface,singh:2020:3d_med_review}, the size and complexity of the related CNN models increase  and thus the need for efficient parallel training methods of these models.
Here, we propose a very specific model parallel training approach for large 2D or 3D CNN models which is loosely inspired by two-level domain decomposition methods~\cite{toselli,QuarteroniValli2008}.
 Similarly as already proposed in~\cite{GuCai:2022:dd_transfer}, instead of considering a \textit{global} CNN, we suggest to use a finite number of \textit{local} CNNs of smaller sizes, acting on smaller parts of the input data. Often, this can be viewed as a decomposition of the original, global CNN into a smaller number of local CNNs but in general, this does not has to be the case since the composition of local CNNs can result in a much more general global neural network.
However, in contrast to~\cite{GuCai:2022:dd_transfer} where the weights of the local CNNs are used as an initialization for the subsequently trained global CNN, here, we do not train the entire global CNN at all. Instead, we train an additional dense feedforward neural network (DNN) which, illustratively spoken, evaluates the local decisions of the local CNNs and creates a final, global decision. We refer to this DNN as global coarse net. To summarize, in contrast to~\cite{GuCai:2022:dd_transfer}, we suggest a completely new CNN-DNN architecture which naturally supports parallel training and which is based on a decomposition of the input data.
Let us note that, for the remainder of this paper, we restrict ourselves to discrete classification problems and CNN models. However, the described approach can also be applied to other network models, e.g., residual neural networks, and different image-based machine learning tasks, e.g., image-based single value regression problems.
 For example, for image-based  parameter estimation problems, our approach can be used to train local subnetworks that generate parameter estimations based on local information only. Then, in a second step, one could train a global coarse net to aggregate the local predictions of the parameter into a final, global parameter estimation.

\subsection{Training of Local Subnetworks on Local Parts of the Input Data}
\label{sec:algo_dd}

We assume that we have a given 2D CNN with a finite number of stacks of convolutional layers and, followingly, a finite number of fully connected layers as described in~\cref{sec:classic_cnn}.
Additionally, we assume that the CNN model takes as input a 2D pixel image with $H\times W$ pixels and outputs a probability distribution with respect to $K \in \mathbb{N}$ classes. In analogy of our approach to domain decomposition methods, the input pixel image and the related CNN correspond to our domain $\Omega$. 
In order to define a parallel training approach for our model, we now decompose the input data in form of 2D pixel images into a finite number $N \in \mathbb{N}$ of sub-images. Note that for input images with 3 channels of $H\times W$ pixels, that is, RGB images, we exclusively decompose the images in the first two dimensions, the height and the width, but not in the third dimension. Hence, each image is decomposed into $N$ sub-images with height $H_i$ and width $W_i, \ i=1, \ldots, N$. Then, for each of these sub-images, we construct corresponding subnetworks,  that is, local CNNs that only operate on a specific part of all input data. For the special case of a decomposition of rectangular 2D images as input data into nonoverlapping, rectangular sub-images, this can be seen as a decomposition of a larger, global CNN into subnetworks by decomposing the width, that is, the channel dimensions of the original network and keeping the depth, that is, the number of layers, constant.
 This means that the subnetworks and the original network always share the same structure but differ in the number of channels of the feature maps, the number of neurons within the fully connected layers, as well as in the number of input nodes. In particular, each subnetwork has as input data only a local part of the original pixel images and is trained exclusively using the sub-images instead of using the entire images. Thus, our approach is a model parallel training approach but not, in its strict sense, data parallel. Let us again remark that our approach is not a classical model parallel training approach, since the original network architecture is also modified as described. Given the analogy to domain decomposition approaches~\cite{toselli,QuarteroniValli2008}, the sub-images and the corresponding subnetworks correspond to local subdomains $\Omega_i,\ i=1,\ldots, N$. We hence refer to the original CNN model as our \textit{global} network or global CNN and to the different subnetworks as \textit{local} networks or local CNNs. 
In~\cref{fig:dd_cnn}, we show a decomposition of the global CNN of VGG3 type~\cite{simonyan:2014:VGGnet} from~\cref{fig:cnn} into $N=4$ local subnetworks corresponding to a decomposition of the global input image into $N=4$ sub-images. Note that all local subnetworks can be trained completely independently of each other and therefore, in parallel on different processors without any communication between the different network models. All subnetworks are trained to optimize a local learning task which is to approximate a functional relation between the respective sub-images and the global class labels. 
Given our application of discrete classification problems, we use a softmax activation function in the final layer of all CNN models (cf.~\cref{sec:algo_dd}). Hence, as output of the different $N$ local subnetworks, we obtain $N$ separate probability distributions for the $K$ classes of our classification problem. 
In order to aggregate the different probability distributions of the local subnetworks, we define an additional global network which evaluates the local decisions of all subnetworks; see~\cref{sec:algo_coarse}.

\begin{figure}[t]
\centering
\includegraphics[width=0.42\textwidth]{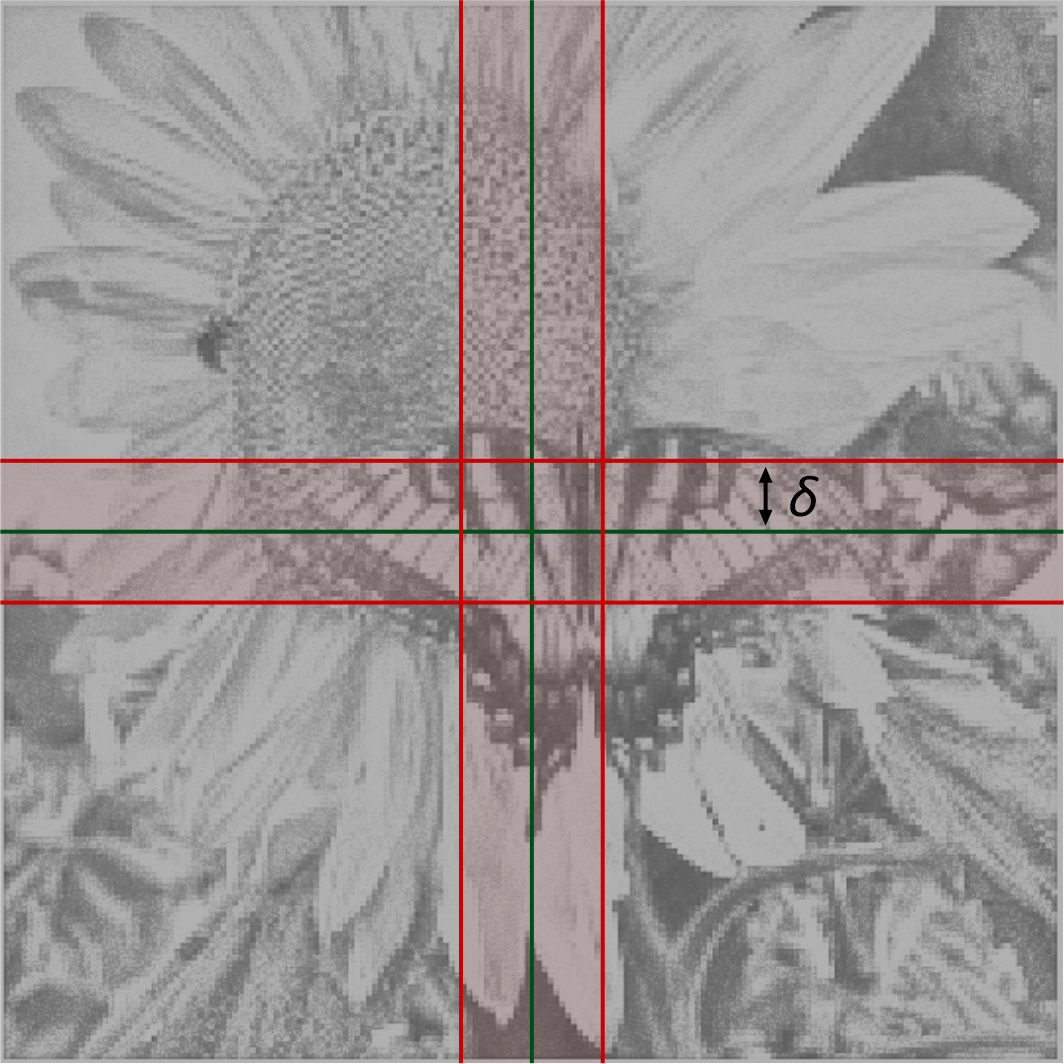}
\hspace{0.5cm}
\includegraphics[width=0.42\textwidth]{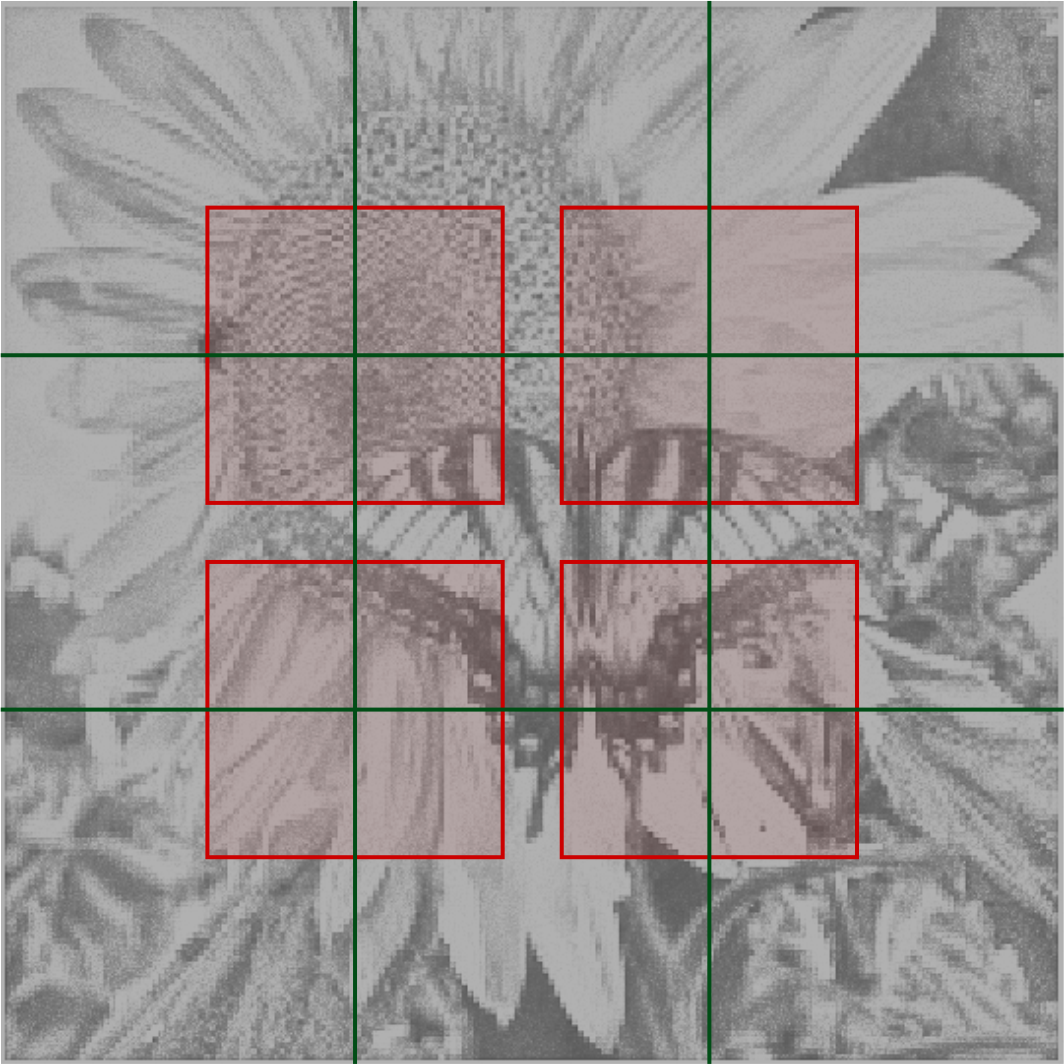}
\caption{\label{fig:dd_nn_ex} Two types of decomposition of the global image into sub-images, inspired by regular domain decompositions~\cite{toselli}. \textbf{Left:} Regular decomposition into $2\times2$ square sub-images with a generic overlap $\delta \geq 0$. We denote decompositions of this type by \textbf{type A}. \textbf{Right:} Decomposition into $3\times 3$ nonoverlapping, square subdomains and additionally, $2\times 2$ square subdomains with a strong overlap.
We denote decompositions of this type by \textbf{type B.} }
\end{figure}

Eventually, let us briefly comment on how we decompose the original images into $N$ local sub-images. With respect to the analogy of overlapping or nonoverlapping domain decomposition approaches~\cite{toselli,QuarteroniValli2008}, it is natural to decompose a given 2D image of $H\times W$ pixels into $p\times p,\ p\in \mathbb{N}$ rectangular sub-images with an overlap $\delta \geq 0$. In~\cref{fig:dd_nn_ex} (left), we show an exemplary decomposition of a square image with $180\times 180$ pixels into $2\times 2=4$ sub-images with a generic overlap $\delta\geq 0$, that is, we consider $p=2$ and $N=4$.
However, also more general decompositions and even irregular decompositions are possible and can be advantageous given that CNNs work on any input data with a tensor product grid-like structure~\cite[Chapt. 9]{Goodfellow:2016:DL}. For a more general decomposition of a square image into square subdomains with a strong overlap, which we also use for our experiments in~\cref{sec:res}, see~\cref{fig:dd_nn_ex} (right). In contrast to overlapping subdomains in classic DDMs, here, the definition of overlapping sub-images does not introduce nor require any communication between the different local neural networks operating on different overlapping sub-images. However, the overlap can help to increase the accuracy of the classification in some test cases since it increases the size of the local sub-images and the amount of global, that is, shared information each local network has.

Finally, note that the described decomposition of a 2D CNN model can also be generalized to 3D CNN models using 3D voxel data as input data. In this case, the global voxel data of size $H\times W \times D$, where $D$ is the depth of the data, can be decomposed in all three dimensions. The corresponding local subnetworks are then obtained from the global network in analogy as for the 2D CNN model except that all of them then use 3D convolutional filters and 3D pooling operations.

\subsection{Definition of a Global Coarse Network to Aggregate the Outputs of the Local Subnetworks}
\label{sec:algo_coarse}

\begin{figure}
\includegraphics[width=0.98\textwidth]{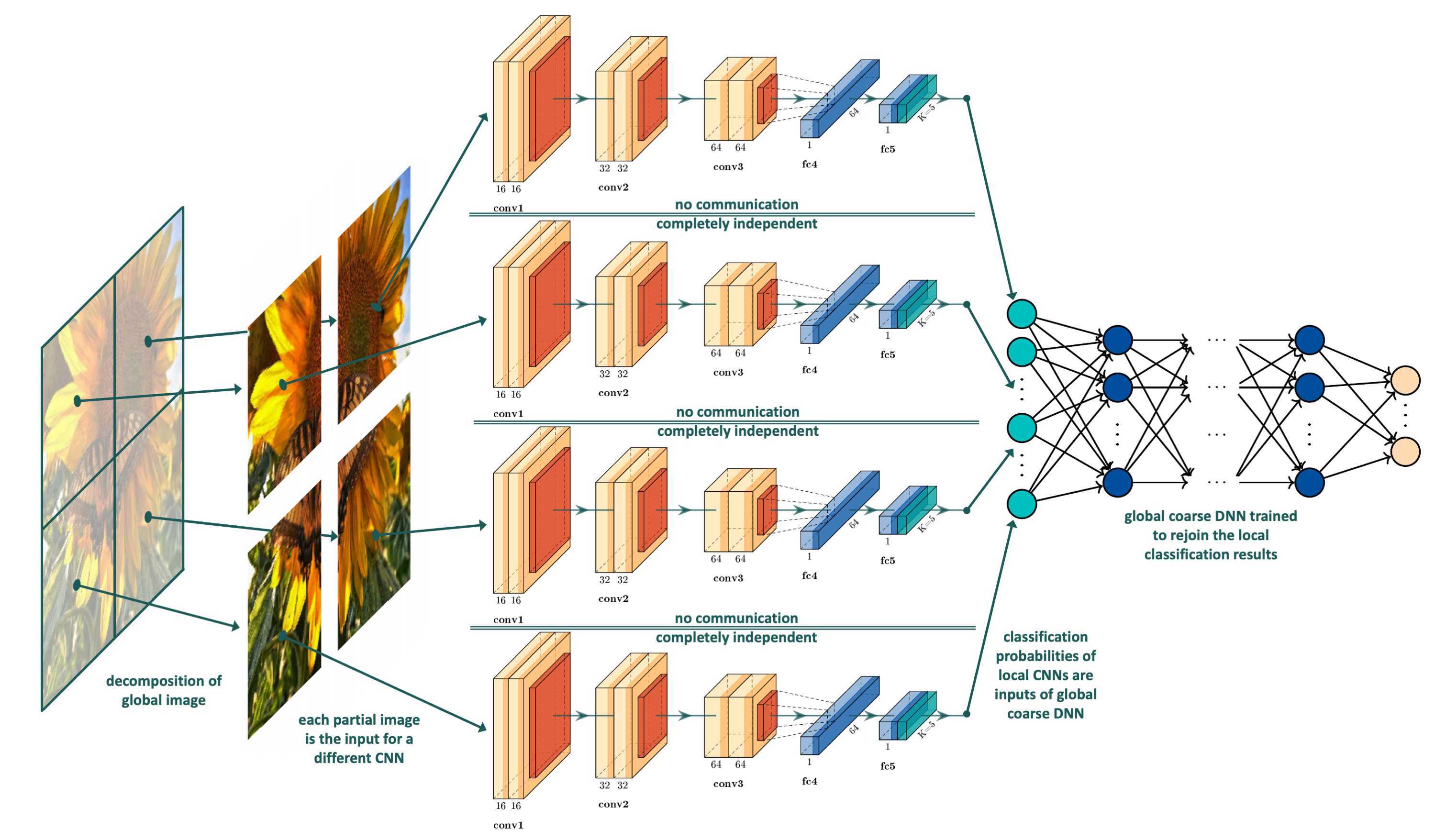}
\caption{Two-level decomposition of a CNN used for an exemplary image recognition problem of the TF-Flowers dataset~\cite{tfflowers}. \textbf{Left:} The original image is decomposed into $p\times p,\ p=2$ nonoverlapping sub-images. \textbf{Middle:} The $p\times p$ sub-images are used as input data for $p\times p$ independent, \textit{local} CNNs. Each of these \textit{local} CNNs has a proportionally smaller size than the \textit{global} CNN from~\cref{fig:cnn} and is exclusively trained on local parts of the original image. \textbf{Right:} The probability values of the local CNNs are used as input data for a DNN to which we refer to as global coarse net. The global coarse net is trained to make a final classification for the decomposed image by weighting the local probability distributions.
See~\cref{fig:cnn} for the color code of the different network layers.}
\label{fig:dd_cnn}
\end{figure}

In this section, we propose an approach for aggregating the different classifications in form of the different class probability distributions of the local subnetworks from~\cref{sec:algo_dd}.
For this purpose, we introduce a DNN model~\cite{Goodfellow:2016:DL}, which has as inputs the different class probabilities of the separate local CNN models. More precisely, if we have a $K$ class classification problem and decompose the global CNN into $N$ subnetworks, the DNN model has $K\ast N$ input nodes. The DNN model is then trained to map the $N$ local probability distributions to the correct classification labels of the original input data of the global network. Thus, the output of this network is a final probability distribution among the $K$ classes of the underlying image classification problem. 
Considering again the analogy to domain decomposition approaches as well as the fact, that the described DNN, illustratively spoken, aggregates the decisions of the local subnetworks, we denote this network by \textit{global coarse net}. Let us note that the global coarse net, so far, is not based on a direct coarse representation of the input images. However, in loose analogy to coarse spaces in domain decomposition, it also has the task to combine and aggregate information from subnetworks that are trained with spatially distant subimages. This would loosely correspond to the role of a coarse space in two-level DDMs such that it also accelerates the distribution of information between distant subdomains. 
In~\cref{fig:dd_cnn}, we show a global coarse net with a generic number of hidden layers that has as inputs the class probability values of $N=4$ local subnetworks. The specific architectures of the trained global coarse nets are defined in~\cref{sec:res} for the different data sets. In all cases, we either apply the softmax activation function for the output layer with $K$ output nodes in order to obtain a valid probability distribution among the $K$ classes if $K>2$ or the sigmoid activation function in case of a binary classification problem for $K=2$; see~\cite{dunne:1997:softmax,sharma:2017:activation}.

Note again that in practical experiments, the global network from~\cref{sec:algo_dd} is not trained at all. 
Instead of training a global CNN for the entire images in our dataset, we decompose the images into local sub-images and train local subnetworks for the different parts of the images. This can be done completely independently on different processors or GPUs of a parallel computer. Subsequently, we train one global coarse net in form of a DNN that aggregates the local outputs of the different local subnetworks. In particular, the global coarse net is trained to make a final classification for the decomposed image by weighting the probability distributions as obtained by the local CNNs.
In principle, the idea of aggregating separate, local models into one final, global decision can also be applied to different and more general machine learning tasks. However, we focus on our application of image recognition problems with convolutional neural networks for the remainder of this paper.

\subsection{Combination with Data Parallel Approaches}

As already mentioned before, we consider our method as a model parallel training approach due to the fact that we decompose the given network architecture into smaller networks. Nonetheless, we also decompose the input images into smaller sub-images and, accordingly to the distribution of the smaller networks to the parallel resources (GPUs), we also distribute the sub-images to the different GPUs. More precisely, we decompose each image into smaller sub-images and each of the sub-images is handled by a different GPU. This could also be interpreted as a data parallel approach but it differs from classical data parallel methods where the dataset is split into different packages but each individual data point is not decomposed or split. In data parallel methods, for each package of the data a network is trained individually and in parallel. All networks should have the same architecture and from time to time the model parameters are communicated and, for example, averaged to obtain a final single model. Consequently, while classical data parallel methods are designed to parallelize training for large data sets with a large number of (small) data points, as, e.g., ImageNet, our approach will be beneficial if the individual data points are large, as, e.g., high-resolution images or 3D data sets from, for example, computer tomographies. For large data sets of individually large images both approaches can be combined easily by first splitting the dataset into several packages and subsequently splitting the individual data points into subdomains. Then, for each package, the same CNN-DNN network is trained on an individual group of GPUs each using a classical data parallel approach. This combination could enable the efficient use of a large number of GPUs. Nonetheless, the averaging routines within the data parallel training approach have to be carefully designed first for the training of the local CNNs and eventually also for the global coarse DNN. Therefore, we postpone an implementation and detailed discussion of this approach to future work.

\section{Datasets and Network Architectures}
\label{sec:data_nets}

In this section, we describe the different datasets that we use to test our proposed approach for different image classification or recognition tasks. In particular, we use three different two-dimensional image datasets, that is, pixel data, and  one three-dimensional image dataset corresponding to CT data, that is, voxel data.  
For all datasets, we have optimized the hyperparameters batch size, initial learning rate, and dropout rate for the global CNN model with a grid search and have used the optimal choice of parameters also for the local CNNs. 
 Let us note that we deliberately chose this procedure to provide a fair comparison between the global model and the proposed parallel training approach. However, apart from benchmarking and comparison purposes, determining the optimal parameters of the global CNN model can be computationally expensive. Hence, we propose the following procedure. First, we suggest to tune the parameters of the local networks. This can be done in parallel for all local networks and also independently from the optimal parameters of the global CNN. For a mathematical description of the computational efforts, we denote by $K$ the number of parameters of the global CNN, by $k$ the number of parameters of the DNN, by $M_{CNN}$ and $M_{DNN}$ the number of points in the search space of the gridsearch with respect to the CNN or the DNN, respectively, and we decompose a two-dimensional image into $p\times p$ sub-images. Then, the costs of a global gridsearch for the global CNN can roughly be estimated at $O(K\ast M_{CNN})$.
In contrast, the costs of tuning the hyperparameters for one local CNN model can roughly be estimated at $O(K/{p^2}\ast M_{CNN})$.
Subsequently, after locally tuning the parameters for each of the local CNN models, we suggest to tune the parameters of the DNN model using the probability distributions from the optimized local CNNs as input data. This requires $O(k \ast M_{DNN})$ iterations such that the overall computational cost of the proposed strategy can be estimated at $O(K/{p^2}\ast M_{CNN})+O(k \ast M_{DNN})$. Given that for all tested datasets, the DNN is relatively shallow, we have $k<<K$ and the computational costs for tuning the DNN are negligible. Therefore, the cost of tuning all local CNNs in parallel and subsequently the DNN is expected to be much cheaper than tuning one global CNN.

\subsection{CIFAR-10 Dataset}
\label{sec:data_nets_cifar}

First, we test the proposed method on the \textit{CIFAR-10} dataset~\cite{Cifar10_TR}.
This dataset consists of $50\,000$ training and $10\,000$ validation images in $K=10$ different classes. The 10 different classes represent airplanes, cars, birds, cats, deers, dogs, frogs, horses, ships, and trucks and there are $6\,000$ images of each class. All the images are RGB images of size $32\times 32$ pixels; see also~\cref{fig:cifar10_ex} (left) for some exemplary images taken from this dataset.

To train a global machine learning model to recognize the 10 different classes of the CIFAR-10 dataset, we use a two-dimensional CNN~\cite{lecun:1989:CNN}. Since the images in the CIFAR-10 dataset have a low resolution, we can obtain relatively high accuracy values of the classification for relatively simple, i.e., small CNNs. In particular, we have chosen a VGG net-type architecture~\cite{simonyan:2014:VGGnet} for our image classification problem since these networks have performed well on various image recognition tasks~\cite{singh:2020:3d_med_review}.
The network, to which we refer to as \textit{VGG3} in the following, consists of 3 stacks of convolutional layers (conv) and two fully connected layers (fc). Each of the stacks of convolutional layers consists of $3\times3$ convolutional filters, uses the ReLU (Rectified Linear Units)~\cite{Nair:2010:RLU} activation function and is followed by a maximum pooling layer and a dropout layer (20\% dropout). In particular, the first stack consists of two convolutional layers with convolutions on the input images of size $\{32,32\}$, followed by the second and the third layer with convolutions on the preceding feature maps of sizes $\{64,64\}$ or $\{128,128\}$, respectively; see also~\cref{fig:cnn} for a visualization of the described convolutional layers. 
For the fully connected layers, we have chosen one layer with 128 neurons and a second layer with $K=10$ neurons followed by a softmax layer.
For the training of the global network, we use an SGD (stochastic gradient descent) method with the Adam (Adaptive Moments) optimizer~\cite{Kingma:2014:MSO} and a batch size of $32$. As the loss function, we choose the categorical crossentropy loss function~\cite{dunne:1997:softmax}. 

Since the images of the CIFAR-10 dataset have a small size of $32\times32$ pixels, we exclusively decompose the images into $2\times 2$ sub-images and thus, always train four local subnetworks for the classification of the sub-images in~\cref{sec:res-cifar10}. Additionally, we consider different values for an overlap $\delta \geq 0$ between the different sub-images.
All the local subnetworks have the same structure as the described global VGG3 network, but with a proportionally smaller size of the convolutional and fully connected layers, corresponding to the smaller size of the sub-images. We train the local subnetworks using the Adam optimizer with a batch size of $32$ and using the categorical crossentropy loss function. 

Finally, for the global coarse net, we choose a dense feedforward neural network with four hidden layers with $\{128, 64, 32, 10 \}$ neurons for the respective layers. For each hidden layer, we use the ReLU activation function and for the output layer, we use the softmax activation function. The global coarse net is also trained using an SGD method with the Adam optimizer~\cite{Kingma:2014:MSO} and a batch size of $32$. As the loss function, we choose again the categorical crossentropy loss function. 

\subsection{TF-Flowers Dataset}
\label{sec:data_nets_flowers}

\begin{figure}
\centering
\includegraphics[width=0.46\textwidth]{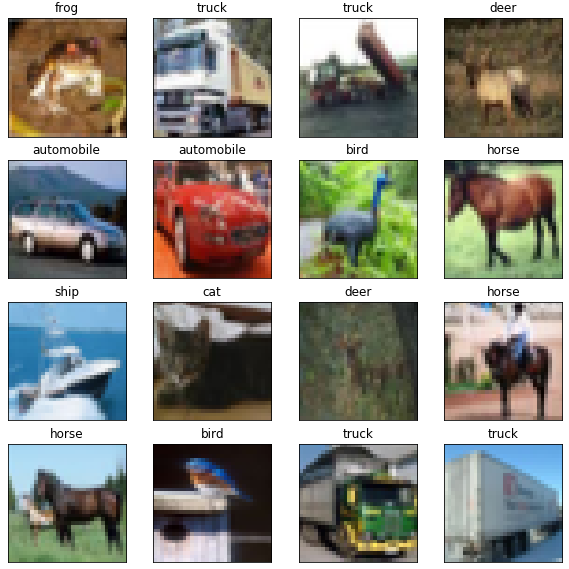}
\hfill
\includegraphics[width=0.46\textwidth]{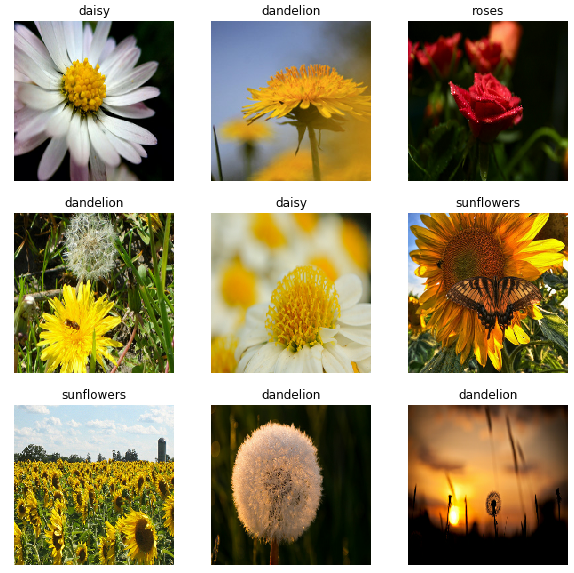}
\caption{\textbf{Left:} Exemplary images of the CIFAR-10 dataset~\cite{Cifar10_TR}. \textbf{Right:} Exemplary images of the TF-Flowers dataset~\cite{tfflowers}.}
\label{fig:cifar10_ex}
\end{figure}

As the second two-dimensional image dataset, we use the \textit{TF-Flowers} dataset~\cite{tfflowers} available from TensorFlow~\cite{tensorflow2015-whitepaper}. This dataset consists of $3\,670$ RGB images of $K=5$ different classes of flowers, that is, daisy, dandelion, roses, sunflowers, and tulips; see~\cref{fig:cifar10_ex} (right) for some exemplary images. We resize all images such that they have an equal size of $180\times 180$ pixels and split the entire dataset into $2\,936$ training and $734$ validation data.

For the image recognition problem trained on the entire images, we train a global 2D convolutional neural network of the same VGG3 type~\cite{simonyan:2014:VGGnet} as for the CIFAR-10 dataset; see~\cref{sec:data_nets_cifar}.

Given that the images have a larger size of $180\times 180$ pixels than the CIFAR-10 dataset, we decompose these images into $p\times p,\ p \in \{2,3,4 \},$ sub-images. Thus, we train four, nine, and 16 local subnetworks, respectively. Again, the local networks have the same structure as the corresponding global network but have a proportionally smaller size of the convolutional and fully connected layers corresponding to the reduced size of the local sub-images in relation to the global image.

To define the global coarse net model, we train a dense feedforward neural network with four hidden layers with $\{128, 64, 32, 5 \}$ neurons for the respective layers, analogously to the CIFAR-10 dataset. For all hidden layers, we use the ReLU activation function and for the output layer, we use the softmax activation function. 
 In~\cref{tab:gridsearch_flowers}, we provide comparative results for different architectures of the global coarse net model with respect to the training and validation accuracy. For the following detailed experiments in~\cref{sec:res-flowers}, we chose the global coarse net with the highest average accuracy with respect to the validation data. 
All described network models, that is, the global network, the $p\times p$ local networks, and the global coarse net are trained using the Adam optimizer~\cite{Kingma:2014:MSO} with a batch size of $32$ and minimizing the categorical crossentropy loss function. 

\begin{table}[th]
\scalebox{0.94}{
\centering
\begin{tabular}{l|l||cccc}
 {\bf Decomp.}  &  & \multicolumn{4}{c}{\bf neurons in hidden layers}  \\
   & & $\{50,30 \}$ & $\{100,50,30 \}$ & $\{128, 64, 32, 5 \}$ & $\{300,256,128,64,32 \}$ \\\hline
 $2\times2$ & {\bf train} & 0.6776 & 0.7676 & 0.8827 & 0.9003 \\
  $\delta=0$  & {\bf val} & 0.5843  & 0.7253 & \bf 0.8154 & 0.7882 \\\hline
   $3\times3$ & {\bf train} & 0.6538 & 0.7889 & 0.8971 & 0.9323 \\
  $\delta=0$  & {\bf val} &  0.5465 & 0.7472 & \bf 0.8738 & 0.7844 \\\hline
   $4\times4$ & {\bf train} & 0.6565 & 0.7656 & 0.8872 & 0.8998 \\
  $\delta=0$  & {\bf val} & 0.5678 & 0.7202 & \bf 0.8589 & 0.8334 \\\hline

\end{tabular}
}
\caption{\label{tab:gridsearch_flowers} Classification accuracy values for the validation and the training data  for the final classification obtained by the CNN-DNN approach for different architectures of the global coarse net for the TF-Flowers dataset~\cite{tfflowers} for the VGG9 architecture and different decompositions of the images of type A (decomp.); cf.~\cref{tab:res_acc_flowers_VGG9}. Note that more combinations of neurons per hidden layers were tested using a Bayesian gridsearch, but, due to space limitations, exclusively a limited number of representative examples are shown.}
\end{table}

\subsection{Extended Yale Face Database B}
\label{sec:data_nets_faces}

\begin{figure}
\centering
\includegraphics[width=0.7\textwidth]{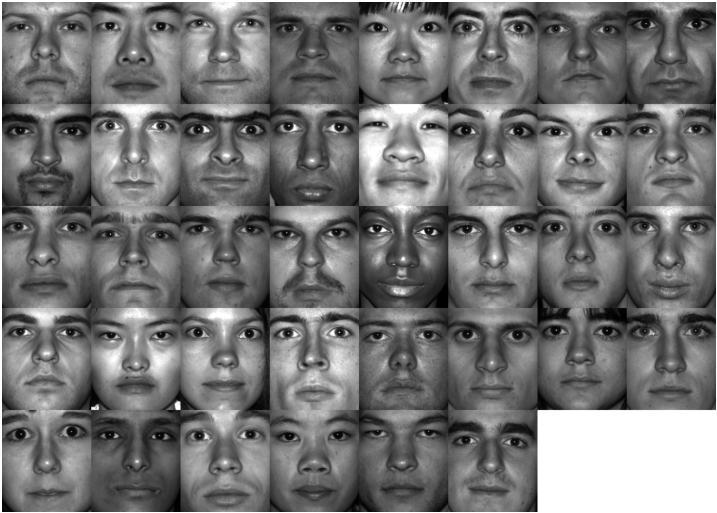}
\caption{One exemplary image for each individual in the Extended Yale Face Database B~\cite{Yale_faces1,Yale_faces2}.}
\label{fig:yale_faces_ex}
\end{figure}

To get an impression of how well our proposed method performs for more complex datasets, we additionally test it for the \textit{Extended Yale Face Database B}~\cite{Yale_faces1,Yale_faces2}, that is, a real-world face recognition problem. This dataset consists of cropped and aligned images of $38$ individuals under nine poses and $64$ different lighting conditions. All images are gray-scale images with a size of $192\times168$ pixels; see~\cref{fig:yale_faces_ex} for an exemplary image of each person.
Given that not all lighting conditions exist for all individuals, we obtain a dataset with a total of $2\,410$ images that we split into $1\,687$ training and $723$ validation data.
Let us note that the task of face recognition is, in general, relatively hard and requires the availability of very large quantities of training data and/or the training of relatively complex network architectures~\cite{parkhi:2015:deepface}.
Thus, as well as due to the relatively small size of the Extended Yale Face Database B, we obtain a significant variance in the related accuracy values of both the global network and the local networks in~\cref{sec:re-faces}. In particular, the results vary significantly for different initializations of the network weights as well as for different splits of the dataset into training and validation data. 
However, the goal of this work is not to train a global machine learning model with the highest possible accuracy value, but to compare the performance of our proposed two-level classification model with the training of a global CNN model. Hence, we believe that the shown results in~\cref{sec:re-faces} for this dataset still deliver a good impression of how well our training approach performs in comparison to the global training.
Analogously to the CIFAR-10 and the TF-Flowers dataset, we train 2D convolutional neural networks for both, the global network model and the local network models. The global network of VGG3 type has the same hyperparameters as for the previous two datasets except for the adjusted size of the output layer corresponding to the number of $K=38$ individuals. 
The local networks have a proportionally smaller size of the convolutional and fully connected layers and we decompose the images into $p\times p,\ p \in \{3,4 \},$ local sub-images. 
For the global coarse net, we use again a dense feedforward neural network with the same number of neurons for each hidden layer as before but with an additional dropout layer (dropout rate of $20\%$) after each hidden layer. 
All the networks are trained using the Adam optimizer with a batch size of $16$.

\subsection{Chest CT Scans}
\label{sec:data_nets_CT}

\begin{figure}[t]
\centering
\includegraphics[width=0.8\textwidth]{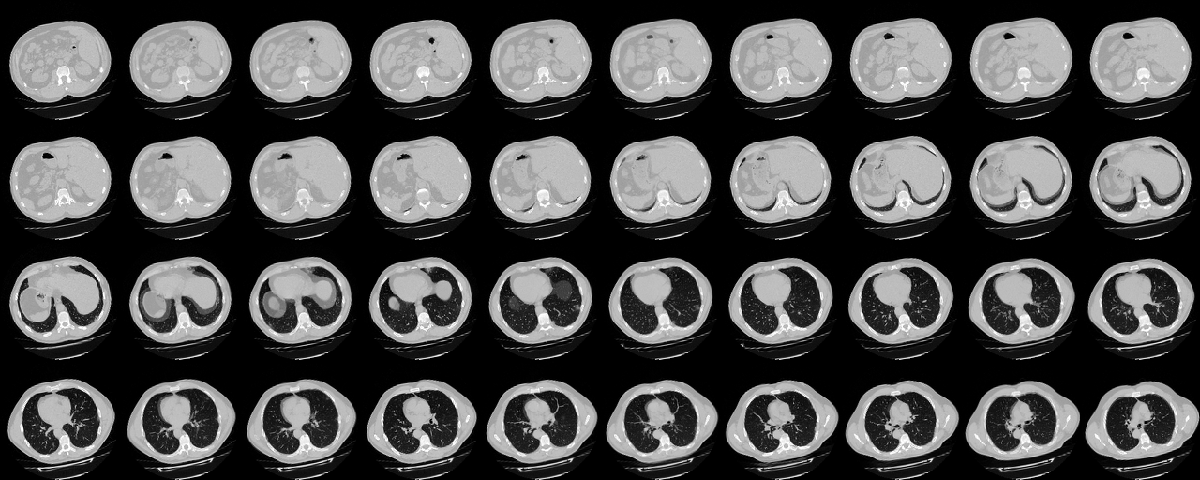}
\caption{Exemplary slices for one chest CT scan taken from the MosMedData dataset~\cite{Chest_CT}.}
\label{fig:CT_lung_ex}
\end{figure}

Finally, to test our approach for more practically relevant and more complex data, we consider a 3D convolutional neural network model to predict the presence of viral pneumonia in computer tomography (CT) scans.
More precisely, we use a subset of the \textit{MosMedData: Chest CT Scans with COVID-19 Related Findings}~\cite{Chest_CT} dataset available from Keras~\cite{chollet2015keras}.
The dataset consists of $200$ Nifti files in which $64$ \textit{slices} of CT scans are stored for each data point. Each of these CT slices consists of $128\times128$ pixels such that we have volumetric data of $128\times128\times64$ voxels. In~\cref{fig:CT_lung_ex}, we show $40$ exemplary slices of one exemplary CT chest scan.
The considered dataset consists of two classes, that is, lung CT scans with COVID-19 related findings as well as without such findings. We use the associated radiological findings of the CT scans to define a binary classification problem to predict the presence of COVID-19 related pneumonia; cf. also~\cite{Chest_CT}. 
We split the data into $140$ training and $60$ validation data using different random seeds.
Let us note that in order to classify the described 3D dataset, we define a global 3D convolutional network model. Since 3D CNNs generalize 2D CNNs by operating on an input of a 3D volume or a sequence of 2D frames (e.g., slices of a CT scan) and thus learn a representation of volumetric data by training three-dimensional filters, 3D CNNs are usually computationally much more expensive than 2D CNNs. Hence, we believe that our proposed network model which facilitates the parallel training of smaller local networks instead of the training of one large, global network has even more potential for 3D CNNs than for 2D CNNs.

The raw files provided from~\cite{Chest_CT} store raw voxel intensities in Hounsfield units (HU) ranging from $-1024$ to above $2000$. Since voxel identities above $400$ are bones with different radiointensities, it is common practice to normalize the CT scans such that they range between $-1000$ and $400$ meaning that we set $400$ HU as the maximum value; see, e.g.,~\cite{zunair:2020:uniformCT}. Additionally, we normalize the voxel values such that they range between $[0,1]$ in a preprocessing phase. 

To define a global network model to classify the slices of the chest CT data, we define a global 3D CNN based on the network architecture in~\cite{zunair:2020:uniformCT}.
Similar as for the previously described 2D CNN models, we build our CNN with different stacks of convolutional and fully connected layers.
We build our model by using four stacks of convolutional layers. Other than for the 2D CNN models, each of these stacks now consists of $3\times3\times3$, that is, 3D convolutional filters. Each of these stacks uses the ReLU activation function~\cite{Nair:2010:RLU} and is followed by a 3D maximum pooling layer and a batch normalization (BN) layer~\cite{ioffe:2015:batch}.
In particular, the first stack consists of one convolutional layer with convolutions on the input voxels of size $64$, and the second, third, and fourth stack consist of one convolutional layer each of convolutions on the preceding feature maps of sizes $64$, $128$, and $256$, respectively.
The last convolutional layer is followed by a global average pooling layer and a fully connected layer with $256$ neurons as well as a dropout layer ($30\%$ dropout). For the output layer, we implement a fully connected layer with one neuron and a sigmoid activation function given that we consider a binary classification problem~\cite{sharma:2017:activation}.
For the training of the global network, we use the Adam optimizer~\cite{Kingma:2014:MSO} with an initial learning rate of $0.0001$, a batch size of $2$ and an early stopping criterion~\cite{prechelt:1998:early_stop} with respect to the validation accuracy with a patience of $15$ epochs. For the loss function, we choose the binary crossentropy loss function.

For the parallel training of the local subnetworks, we use both, a two- and a three-dimensional decomposition of the volumetric data. For a first test, we decompose the CT slices only in two directions but use all 64 slices for all subnetworks. Hence, we decompose the voxel data into $p\times p\times 1,\ p\in \{2,3,4 \},$ subsets of the voxel data. In the second and more general test case, we decompose the CT slices in all three directions, that is, we decompose the data into $4\times4\times2$ subsets of the voxel data; see also~\cref{sec:res-CT}. 
In all cases, we use a proportionally smaller size for the convolutional layers of the local networks as for the global network.
The global coarse net for this dataset is defined as follows. We define a feedforward neural network with four hidden layers, i.e., $\{ 64,32,16,1\}$ neurons within the hidden layers. Additionally, we use $40\%$ dropout within all hidden layers. For each hidden layer, we use the ReLU activation function and for the output layer, we use the sigmoid activation function. 
For the training of the global coarse net, we use the Adam optimizer with the initial learning rate $0.001$, batch size $2$, and the binary crossentropy loss function.

\begin{table}
\centering
\scalebox{0.97}{
\begin{tabular}{l|l|l|l}
{\bf network} & {\bf $\#$ params} & {\bf MFLOPs}  & {\bf TtEval} \\\hline\hline
\multicolumn{4}{c}{\bf CIFAR-10, $2\times 2$ sub-images} \\\hline
VGG3 global & 0.5506M & \phantom{1}0.7779  & 207ms \\
VGG3 local & 0.0892M & \phantom{1}0.0497  & \phantom{1}82ms \\\hline
VGG9 global & 4.7868M & \phantom{1}1.3404  & 312ms \\
VGG9 local &1.1959M & \phantom{1}0.0851  & \phantom{1}98ms\\\hline
ResNet20 global & 0.2718M & \phantom{1}0.8110  & 623ms  \\
ResNet20 local &0.0691M & \phantom{1}0.0513  & 154ms \\\hline\hline
\multicolumn{4}{c}{\bf TF-Flowers, $3\times 3$ sub-images} \\\hline
VGG3 global & 8.2178M & 24.6064  & 298ms \\
VGG3 local & 0.2732M & \phantom{1}0.6987  & \phantom{1}97ms \\\hline
VGG9 global &6.3593M & 41.6092 &  443ms \\
VGG9 local &1.3266M & \phantom{1}1.0736 & 153ms \\\hline
ResNet20 global &0.2715M &25.6608  & 787ms  \\
ResNet20 local &0.0689M & \phantom{1}0.7206  & 302ms \\\hline\hline
\multicolumn{4}{c}{\bf Extended Yale Face Database B, $4\times 4$ sub-images} \\\hline
VGG3 global & 8.5499M & 24.1328  &256ms \\
VGG3 local & 0.1972M & \phantom{1}0.3733  & \phantom{1}85ms \\\hline\hline
\multicolumn{4}{c}{\bf Chest CT scans, $2\times 2\times1$ sub-images} \\\hline
VGG global &1.3529M &11.9684  & 403ms \\
VGG local &0.3396M & \phantom{1}0.8069  & 127ms \\\hline\hline

\end{tabular}
}
\caption{\label{tab:params_flops} Number of trainable model parameters ($\#$ params), FLOPS, and average time for evaluation (TtEval) of the different global network models and one of their corresponding local network models for the CIFAR-10, the TF-Flowers, the Extended Yale Face data, and the chest CT scan data for the given image decompositions. Here, 'M' refers to $10^6$, i.e., one million.
}
\end{table}

\section{Experimental Results}
\label{sec:res}

In this section, the proposed CNN-DNN architecture is evaluated by different experiments for different image classification or recognition problems. In all cases, we train CNNs of different structures and architectures; see also~\cref{sec:data_nets}. In particular, we train 2D CNNs for two image classification datasets, that is, the CIFAR-10~\cite{Cifar10_TR} and the TF-Flowers~\cite{tfflowers} dataset, as well as for the face recognition problem with respect to the Extended Yale Face Database B~\cite{Yale_faces1,Yale_faces2}, and we train a 3D CNN for the chest CT scans data~\cite{Chest_CT}.
 In~\cref{tab:params_flops}, we summarize the number of model parameters, FLOPs, and average time to evaluation for the global networks and some exemplary local networks for the four tested datasets. Let us note that, in~\cref{tab:params_flops}, the reported FLOPs of a network refer to the FLOPs of one forward propagation of the network with the batch size of $1$; see also~\cite[sect. A.1]{molchanov2016pruning} for a formula of the FLOPs computation for convolutional and feedfoward layers in a neural network. For the FLOPs computations, we have used the tensorflow/keras model profiler~\cite{model_prof}.
All experiments have been performed on a workstation with $8$ NVIDIA Tesla V100 32GB GPUs using python 3.6 and TensorFlow 2.4~\cite{tensorflow2015-whitepaper}.

We compare the classification accuracies of both, the training and the validation data of the respective global CNN and the proposed decomposed CNN-DNN approach when evaluating the trained global coarse net for all four tested datasets. Additionally, for the TF-Flowers~\cite{tfflowers} and the chest CT scan data~\cite{Chest_CT}, we compare the runtimes on the aforementioned GPU cluster needed for the training of the respective networks.
Let us note that since, at this point, we do not have a sophisticated parallel implementation of the two-level decomposition CNN-DNN model available, with respect to the training times, we report the time required for the training of the slowest of the $N$ local networks which are trained completely independently of each other. Additionally, we report the training time required for the global coarse net. The sum of both values is representative for the entire training time of our proposed CNN-DNN approach. This sum is then compared to the time necessary for the training of the global CNN. All networks are trained on one GPU of the aforementioned cluster and are, so far, not parallelized such that the training of a network can be assigned to multiple GPUs.

Additionally, note that the aim of our work is not to define a new neural network model that results in the highest possible accuracy values for the different image classification problems. Instead, we are interested in comparing the performance of a given global CNN model to our proposed two-level decomposition CNN-DNN approach in terms of accuracy, generalization properties, and the required training times.
For the four different tested datasets, we will see that the decomposed CNN-DNN approach, which is training local, proportionally smaller CNNs in parallel and subsequently training a global coarse net, in the best cases often leads to even higher classification accuracies than training a single, proportionally larger global CNN. Moreover, the total time to obtain a final classification for the input data can be drastically reduced using our decomposed training strategy.

\subsection{Experiments on CIFAR-10}
\label{sec:res-cifar10}

At first, we test our approach from~\cref{sec:algo} for the CIFAR-10 dataset~\cite{Cifar10_TR}.
A detailed description of this dataset and the trained network architectures are given in~\cref{sec:data_nets_cifar}. Note again that the images from the CIFAR-10 dataset are relatively small with only $38\times 38$ pixels. Hence, we decompose the images in this section exclusively into $2 \times 2$ nonoverlapping or overlapping square sub-images of type A; see~\cref{fig:dd_nn_ex} (left).

In~\cref{tab:res_acc_cifar}, we summarize the training and validation accuracies of the global CNN model of VGG3 type as well as of the local, decomposed CNNs and the final CNN-DNN using the additional global coarse net. For all three tested values of the overlap $\delta$, the validation accuracy of the CNN-DNN is comparable to the respective results of the global CNN and even slightly higher for the overlap $\delta=4$ pixels. Moreover, we can observe that, when simply evaluating the local CNNs for the respective sub-images, we obtain remarkably lower training and validation accuracies as for the global CNN. Both the local accuracy values as well as the classification accuracies for the global coarse net increase with an increasing overlap $\delta$ between the four sub-images.
 This implies that a stronger coupling between the different sub-images in form of a larger overlap can help to increase the performance of the CNN-DNN.

\begin{table}[th]
\centering
\scalebox{0.94}{
\begin{tabular}{l||c||c|c|c||c}
 {\bf Decomp.} & {\bf global CNN} & \multicolumn{3}{c||}{\bf $p^2$ local CNN} & {\bf CNN-DNN} \\\hline 
  & & avg & min & max & \\\hline
type A & 0.7818 & 0.5091 & 0.4867 & 0.5352 & 0.7669 \\
  $2\times 2$, $\delta=0$ & (0.7821) & (0.5381) & (0.5164) & (0.5605) & (0.8071) \\\hline
type A & 0.7818 & 0.5620 & 0.5353 & 0.5862  & 0.7808 \\
 $2\times 2$, $\delta=2$ & (0.7821) & (0.6063) & (0.5786) & (0.6390) & (0.8343) \\\hline
type A & 0.7818 & 0.5983 & 0.5785 & 0.6149 & 0.7973 \\
$2\times 2$, $\delta=4$ & (0.7821) & (0.6053) & (0.4760) & (0.6629) & (0.8448) \\\hline
\end{tabular}
}
\caption{\label{tab:res_acc_cifar} Classification accuracy values for the validation and the training data (in brackets) for the global CNN and the proposed parallel training approach, i.e., the local CNNs and the final classification obtained by the CNN-DNN approach for the CIFAR-10 dataset~\cite{Cifar10_TR} and different decompositions of the images of type A. See~\cref{sec:data_nets_cifar} for a detailed description of the dataset and the trained 2D CNN architectures.}
\end{table}

\begin{table}[th]
\centering
\scalebox{0.94}{
\begin{tabular}{l||c||c|c|c||c}
 {\bf Decomp.} & {\bf global CNN} & \multicolumn{3}{c||}{\bf $p^2$ local CNN} & {\bf CNN-DNN} \\\hline 
  & & avg & min & max & \\\hline
type A & 0.7585 & 0.4862 & 0.4277 & 0.5173 & 0.7999 \\
  $2\times 2$, $\delta=0$ & (0.8487) & (0.5122) & (0.4838) & (0.5553) & (0.8663) \\\hline
type A & 0.7585 & 0.5545 & 0.5121 & 0.5677  & 0.8008 \\
 $2\times 2$, $\delta=2$ & (0.8487) & (0.5899) & (0.5747) & (0.6007) & (0.8676) \\\hline
type A & 0.7585 & 0.6121 & 0.5704 & 0.6336 & 0.8226 \\
$2\times 2$, $\delta=4$ & (0.8487) & (0.6434) & (0.5464) & (0.6788) & (0.8887) \\\hline
\end{tabular}
}
\caption{\label{tab:res_acc_cifar_VGG9}  Classification accuracy values for the validation and the training data (in brackets) for the global CNN and the proposed parallel training approach, i.e., the local CNNs and the final classification obtained by the CNN-DNN approach for the CIFAR-10 dataset~\cite{Cifar10_TR}, the VGG9 and different decompositions of the images of type A. See~\cref{sec:data_nets_cifar} for a detailed description of the dataset and the trained 2D CNN architectures.}
\end{table}

\begin{table}[th]
\centering
\scalebox{0.94}{
\begin{tabular}{l||c||c|c|c||c}
 {\bf Decomp.} & {\bf global CNN} & \multicolumn{3}{c||}{\bf $p^2$ local CNN} & {\bf CNN-DNN} \\\hline 
  & & avg & min & max & \\\hline
type A & 0.8622 & 0.5225 & 0.4897 & 0.5334 & 0.8784 \\
  $2\times 2$, $\delta=0$ & (0.9343) & (0.5521) & (0.5002) & (0.5643) & (0.9467) \\\hline
type A & 0.8622 & 0.5303 & 0.4967 &  0.5555 & 0.8829 \\
 $2\times 2$, $\delta=2$ & (0.9343) & (0.5435) & (0.5189) & (0.5688) & (0.9447) \\\hline
type A & 0.8622 & 0.5667 & 0.5121 & 0.5828 & 0.8962 \\
$2\times 2$, $\delta=4$ & (0.9343) & (0.5787) & (0.5333) & (0.5901) & (0.9525) \\\hline
\end{tabular}
}
\caption{\label{tab:res_acc_cifar_ResNet20} Classification accuracy values for the validation and the training data (in brackets) for the global CNN and the proposed parallel training approach, i.e., the local CNNs and the final classification obtained by the CNN-DNN approach for the CIFAR-10 dataset~\cite{Cifar10_TR}, the ResNet20 and different decompositions of the images of type A. See~\cref{sec:data_nets_cifar} for a detailed description of the dataset and the trained 2D CNN architectures.}
\end{table}

\subsection{Experiments on TF-Flowers}
\label{sec:res-flowers}

In this section, we test our proposed CNN-DNN approach for the TF-Flowers dataset~\cite{tfflowers}, that is, RGB images of size $180 \times 180$ pixels.
For a detailed description of this dataset and the chosen VGG net architectures, see~\cref{sec:data_nets_flowers}. 
In~\cref{tab:res_acc_flowers}, we show the obtained accuracy values when evaluating the global CNN, the local networks and the CNN-DNN using the additional global coarse net for both, the training and the validation data for different decompositions of the images. 
In the first three rows of~\cref{tab:res_acc_flowers}, we consider three different nonoverlapping decompositions of the images into $p \times p,\ p \in \{2,3,4\}$ square sub-images (type A, see~\cref{fig:dd_nn_ex} (left)). 
Both the classification accuracies for the training and the validation data of the local CNNs and the global coarse net first increase when comparing $p=2$ to $p=3$ and slightly decrease when comparing $p=3$ to $p=4$. This can imply that the decomposition of the global image into a finite number of sub-images and classifying the sub-images separately can help to identify local features of the images more accurately. On the other hand, the sub-images should not be chosen too small since for the extreme case with a sub-image corresponding to only a few pixels, a unique classification can hardly be defined. 

\begin{figure}
\centering
\includegraphics[width=0.43\textwidth]{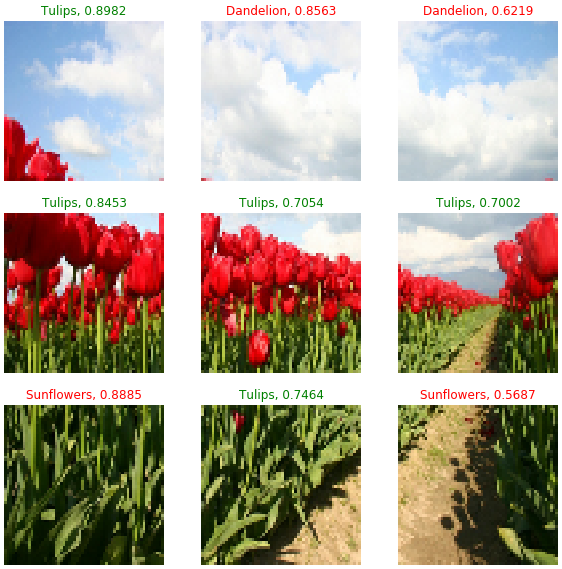}
\hspace{0.5cm}
\includegraphics[width=0.43\textwidth]{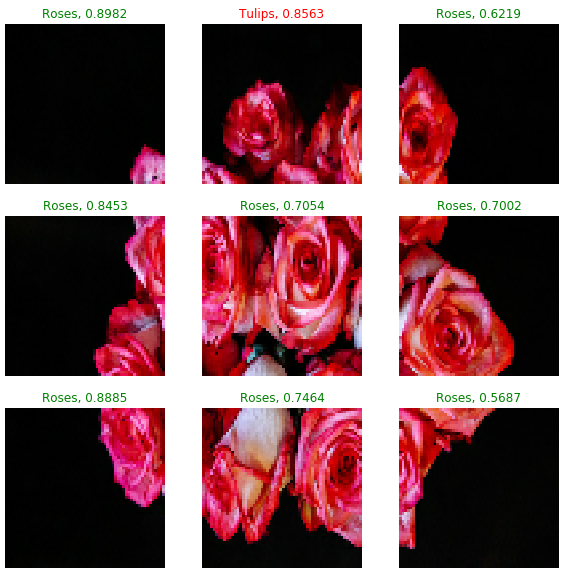}

\vspace{0.9cm}

\includegraphics[width=0.43\textwidth]{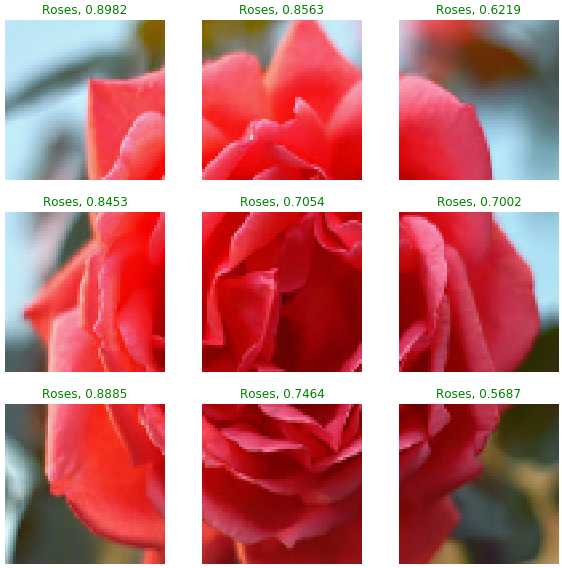}
\hspace{0.5cm}
\includegraphics[width=0.43\textwidth]{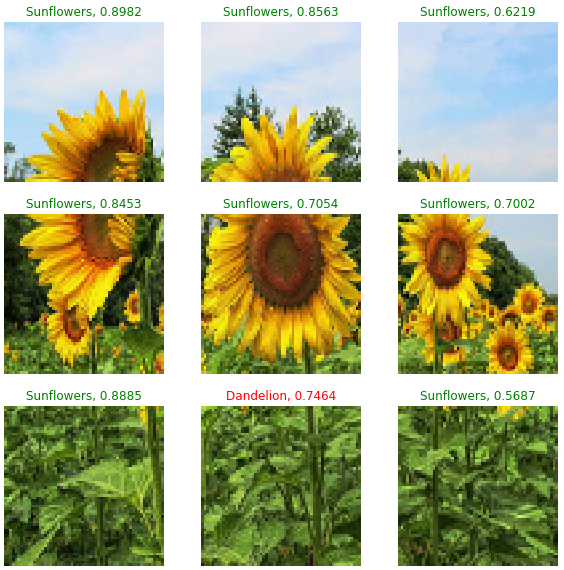}
\caption{Exemplary classification results for the TF-Flowers~\cite{tfflowers} of the local CNNs for the respective local sub-images obtained from a decomposition of the corresponding global image into $3\times3$ nonoverlapping sub-images. Above each local sub-image, we show the obtained classification by the local CNN, either in green, if the classification corresponds to the correct label, or in red, otherwise. Additionally, we show the related class probability, that is, how confident the network is.}
\label{fig:ex_local_nn_flowers}
\end{figure}

Additionally, we observe from~\cref{tab:res_acc_flowers} that with an increasing number of sub-images, the results between the different subnetworks tend to vary stronger such that we obtain a higher range between the minimum and the maximum measured classification accuracies.
In~\cref{fig:ex_local_nn_flowers}, we visualize the obtained classification labels of the different subnetworks for four exemplary images of the dataset which are all decomposed into $3\times 3 $ nonoverlapping sub-images. Hence, we have trained $N=9$ independent subnetworks. As we can see from~\cref{fig:ex_local_nn_flowers}, for three out of the four image examples, not all nine subnetworks classify the respective sub-image to the correct flower class. In particular, we obtain wrong classifications for sub-images where no characteristics of the flowers are included at all, e.g., for sub-images that only contain pixel values related to a blue sky. Moreover, we can observe that the different subnetworks differ in the related class probabilities for the assigned labels, that is, how confident the networks are. 
Both observations emphasize the strong importance and the leading role of the additionally trained global coarse net. In addition to a classical one-level decomposition of the global CNN model, the proposed two-level decomposition trains an additional DNN which enables us to weight the different local classifications of the subnetworks against each other and thus, to increase the final classification accuracy of our proposed training strategy. 
In particular, for all tested decompositions, the CNN-DNN using the global coarse net provides a higher validation accuracy than the global CNN which is a very promising observation. Besides, for the global coarse net, the training and validation accuracy values are fairly similar indicating satisfying generalization properties of the trained network model. 
In the last row of~\cref{tab:res_acc_flowers}, we report additional results for a decomposition of the images of type B (see~\cref{fig:dd_nn_ex} (right)) into $3\times 3$ nonoverlapping, square subdomains and $2\times 2$ additional square subdomains with a strong overlap. 
First, we observe that our approach does also work for the tested decomposition of the global image into sub-images which have a stronger overlap than for decompositions of type A such that the different sub-images overlap not only at the border areas.
Second, we observe that in comparison with the tested decomposition of type A into $3\times 3$ sub-images, the final accuracy of the global coarse net is even increased. This implies that more sophisticated decompositions of the input data are also possible and can be beneficial for our approach. 

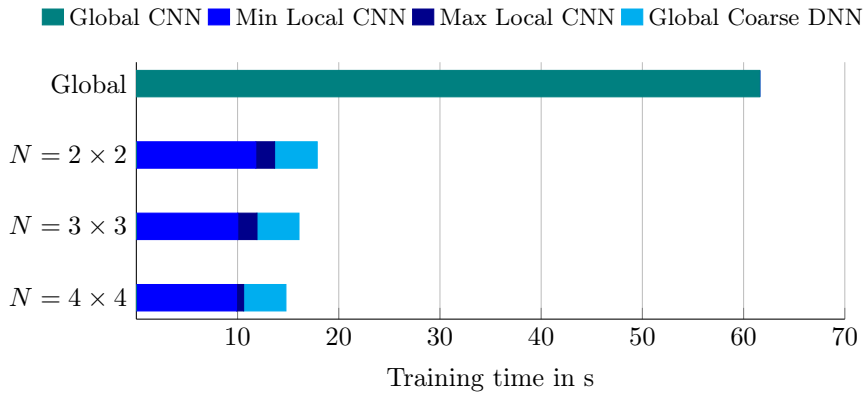
\begin{figure}
\centering
\begin{tikzpicture}
\begin{axis} [xbar stacked,width=11cm,height=5cm, axis y line*=none,
    axis x line*=bottom, ytick={0,1,2,3}, yticklabels={$N=4\times 4$, $N=3\times 3$, $N=2\times 2$, Global}, xlabel = Training time in s, xmajorgrids=true, xmin=0, xmax=70, xtick={10,20,30,40,50,60,70}, enlarge x limits=0.0, legend style={legend columns=4,at={(0.45,1.25)},anchor=north,draw=none, font=\small}]
\addplot+[xbar,teal] coordinates { % global CNN
(0,0)
(0,1)
(0,2)
(61.59,3)
};
\addplot+[xbar,blue] coordinates { % min local CNN
(9.90, 0)
(10.02, 1)
(11.80, 2)
(0, 3)
};
\addplot+[xbar,darkblue] coordinates { % max local CNN -- just difference max-min
(0.70, 0)
(1.90, 1)
(1.86, 2)
(0, 3)
};
\addplot+[xbar,cyan] coordinates { % global coarse CNN
(4.18, 0)
(4.15, 1)
(4.21, 2)
(0, 3)
};
\legend{Global CNN, Min Local CNN, Max Local CNN, Global Coarse DNN}
\end{axis}
\end{tikzpicture}
\caption{Comparison of the training times for the global CNN models and our proposed decomposed approach for the TF-Flowers dataset~\cite{tfflowers}. We show in teal the training time required for the global 2D CNN model using the entire TF-Flowers images as input data. In dark and light blue, we show the maximum required training time for one of the local CNN models and the required training time for the global coarse net, respectively, for different decompositions of the data and the CNN. See~\cref{tab:res_acc_flowers} for the corresponding classification accuracies of the trained models. }	
\label{fig:runtime_flowers}
\end{figure}

Finally, in~\cref{fig:runtime_flowers}, we provide the training times for the global CNN, the local subnetworks and the CNN-DNN using the global coarse net for different decompositions of the images into sub-images and hence, different numbers of subnetworks using the same computing resources.  Given that, for all experiments, we have used a GPU cluster with $8$ NVIDIA GPUs, we always train the global network on one GPU and equally distribute the training of the local CNNs to the $8$ GPUs. More precisely, the training of the $k$-th local CNN is assigned to GPU with index=$k \mod 8$. Subsequently, we train the global coarse DNN on one GPU. Note that for the decomposed approach, we always provide the maximum training time needed for the training of one of the $N$ local networks which in sum with the training time of the global coarse net can be seen as the total training time for our model parallel approach.
All training times are measured in seconds (s) and we train all networks for $500$ epochs without using early stopping. 
As expected, the maximum training time for the local subnetworks decreases with increasing number of subnetworks due to the proportionally smaller size of the subnetworks and lower dimensional input data. The training time for the global coarse net remains approximately constant for all three tested decompositions. 
In sum, the training time for the global 2D CNN is $3.45$ times larger than for the decomposed approach with $2\times 2$ subnetworks and $4.17$ times larger than for the decomposition into $4\times 4$ subnetworks, respectively. We expect that the required training time can be reduced even more drastically for 3D CNN models operating on three-dimensional input data; see also~\cref{sec:res-CT}.

\begin{table}[th]
\centering
\scalebox{0.94}{
\begin{tabular}{l||c||c|c|c||c}
  {\bf Decomp.} & {\bf global CNN} & \multicolumn{3}{c||}{\bf $p^2$ local CNN} & {\bf CNN-DNN} \\\hline 
  & & avg & min & max & \\\hline
type A & 0.6456 & 0.5192 & 0.4888 & 0.5334 & 0.6938 \\
 $2\times 2$, $\delta=0$ & (0.9255) & (0.5570) & (0.5068) & (0.5831) & (0.7552) \\\hline
type A  & 0.6456 & 0.5892 & 0.5215 & 0.6492 & 0.8621 \\
 $3\times 3$, $\delta=0$ & (0.9255) & (0.6299) & (0.5661) & (0.6948) & (0.8848) \\\hline
type A & 0.6456 & 0.5598 & 0.4955 & 0.6262 & 0.8471  \\
 $4\times 4$, $\delta=0$ & (0.9255) & (0.5882) & (0.5199) & (0.6725) & (0.8593) \\\hline
type B, $3\times 3$ + & 0.6456 & 0.5911 & 0.5178 & 0.6502 & 0.8859  \\
$2\times 2$, $\delta=0$ & (0.9255) & (0.5999) & (0.5223) & (0.6676) & (0.9027) \\\hline

\end{tabular}
}
\caption{\label{tab:res_acc_flowers} Classification accuracy values for the validation and the training data (in brackets) for the global CNN and the proposed parallel training approach, i.e., the local CNNs and the final classification obtained by the CNN-DNN approach for the TF-Flowers dataset~\cite{tfflowers} and different decompositions of the images of type A and type B. See~\cref{sec:data_nets_flowers} for a detailed description of the dataset and the trained 2D CNN architectures.}
\end{table}

\begin{table}[th]
\centering
\scalebox{0.94}{
\begin{tabular}{l||c||c|c|c||c}
  {\bf Decomp.} & {\bf global CNN} & \multicolumn{3}{c||}{\bf $p^2$ local CNN} & {\bf CNN-DNN} \\\hline 
  & & avg & min & max & \\\hline
type A & 0.7887 & 0.6122 & 0.5974 & 0.6802 & 0.8154 \\
 $2\times 2$, $\delta=0$ & (0.9321) & (0.6667) & (0.6124) & (0.6999) & (0.8827)  \\\hline
type A  & 0.7887 & 0.6445 & 0.5878 &  0.6889 & 0.8738 \\
 $3\times 3$, $\delta=0$ & (0.9321) & (0.6973)& (0.6224) & (0.7112) & (0.8971) \\\hline
type A & 0.7887 & 0.6353 & 0.5969 & 0.6878 & 0.8589  \\
 $4\times 4$, $\delta=0$ & (0.9321) & (0.6777) & (0.6202) & (0.7024) & (0.8872) \\\hline
type B, $3\times 3$ + & 0.7887 & 0.6544 & 0.6102 & 0.6872 & 0.8689  \\
$2\times 2$, $\delta=0$ & (0.9321) & (0.6898) & (0.6445) & (0.7089) & (0.9056) \\\hline

\end{tabular}
}
\caption{\label{tab:res_acc_flowers_VGG9} Classification accuracy values for the validation and the training data (in brackets) for the global CNN and the proposed parallel training approach, i.e., the local CNNs and the final classification obtained by the CNN-DNN approach for the TF-Flowers dataset~\cite{tfflowers}, the VGG9 and different decompositions of the images of type A and type B. See~\cref{sec:data_nets_flowers} for a detailed description of the dataset and the trained 2D CNN architectures.} 
\end{table}

\begin{table}[th]
\centering
\scalebox{0.94}{
\begin{tabular}{l||c||c|c|c||c}
  {\bf Decomp.} & {\bf global CNN} & \multicolumn{3}{c||}{\bf $p^2$ local CNN} & {\bf CNN-DNN} \\\hline 
  & & avg & min & max & \\\hline
type A & 0.8227 & 0.6202 & 0.5887 & 0.6781 & 0.8475 \\
 $2\times 2$, $\delta=0$ & (0.9178) & (0.6746) & (0.6274) & (0.6998) & (0.9454)  \\\hline
type A  & 0.8227 & 0.5878 & 0.5632 & 0.6202  & 0.8279 \\
 $3\times 3$, $\delta=0$ & (0.9178) & (0.6444) & (0.5820) & (0.6666) & (0.9101) \\\hline
type A & 0.8227 & 0.5676 & 0.5303 & 0.5887 & 0.8068  \\
 $4\times 4$, $\delta=0$ & (0.9178) & (0.6032) & (0.5555) & (0.6479) & (0.8892) \\\hline
type B, $3\times 3$ + & 0.8227 & 0.5773 & 0.5454 & 0.6389 & 0.8375  \\
$2\times 2$, $\delta=0$ & (0.9178) & (0.6288) & (0.5702) & (0.6738) & (0.9398) \\\hline

\end{tabular}
}
\caption{\label{tab:res_acc_flowers_ResNet20} Classification accuracy values for the validation and the training data (in brackets) for the global CNN and the proposed parallel training approach, i.e., the local CNNs and the final classification obtained by the CNN-DNN approach for the TF-Flowers dataset~\cite{tfflowers}, the ResNet20 and different decompositions of the images of type A and type B. See~\cref{sec:data_nets_flowers} for a detailed description of the dataset and the trained 2D CNN architectures.} 
\end{table}

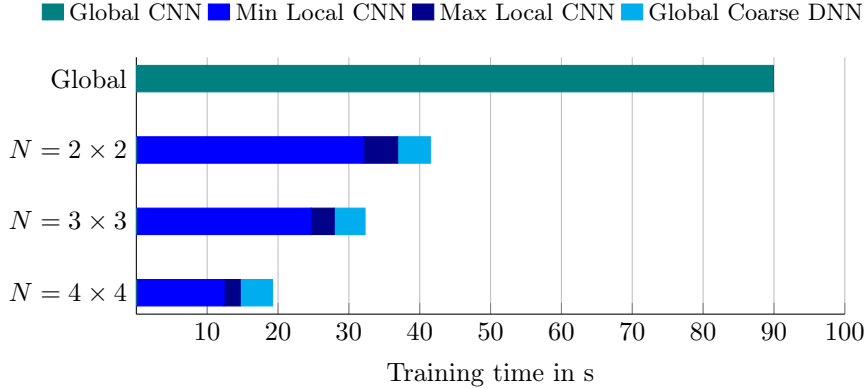
\begin{figure}
\centering
\begin{tikzpicture}
\begin{axis} [xbar stacked,width=11cm,height=5cm, axis y line*=none,
    axis x line*=bottom, ytick={0,1,2,3}, yticklabels={$N=4\times 4$, $N=3\times 3$, $N=2\times 2$, Global}, xlabel = Training time in s, xmajorgrids=true, xmin=0, xmax=100, xtick={10,20,30,40,50,60,70,80,90,100}, enlarge x limits=0.0, legend style={legend columns=4,at={(0.45,1.25)},anchor=north,draw=none, font=\small}]
\addplot+[xbar,teal] coordinates { % global CNN
(0,0)
(0,1)
(0,2)
(89.89,3)
};
\addplot+[xbar,blue] coordinates { % min local CNN
(12.38, 0)
(24.66, 1)
(32.10, 2)
(0, 3)
};
\addplot+[xbar,darkblue] coordinates { % max local CNN -- just difference max-min
(2.28, 0)
(3.27, 1)
(4.79, 2)
(0, 3)
};
\addplot+[xbar,cyan] coordinates { % global coarse CNN
(4.55, 0)
(4.35, 1)
(4.61, 2)
(0, 3)
};
\legend{Global CNN, Min Local CNN, Max Local CNN, Global Coarse DNN}
\end{axis}
\end{tikzpicture}
\caption{Comparison of the training times for the global CNN models and our proposed decomposed approach for the TF-Flowers dataset~\cite{tfflowers} and the VGG9. We show in teal the training time required for the global 2D CNN model using the entire TF-Flowers images as input data. See~\cref{fig:runtime_flowers} for color explanations. 
See~\cref{tab:res_acc_flowers_VGG9} for the corresponding classification accuracies of the trained models. }	
\label{fig:runtime_flowers_VGG9}
\end{figure}

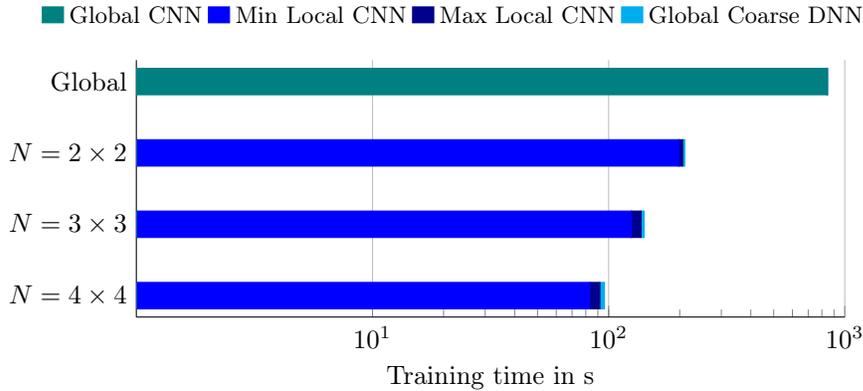
\begin{figure}
\centering
\begin{tikzpicture}
\begin{axis} [xbar stacked,width=11cm,height=5cm, axis y line*=none,
    axis x line*=bottom, ytick={0,1,2,3}, yticklabels={$N=4\times 4$, $N=3\times 3$, $N=2\times 2$, Global}, xlabel = Training time in s, xmajorgrids=true, , xmode= log, xmin=1, xmax=1000, xtick={10,100,1000}, enlarge x limits=0.0, legend style={legend columns=4,at={(0.45,1.25)},anchor=north,draw=none, font=\small}]
\addplot+[xbar,teal] coordinates { % global CNN
(1,0)
(1,1)
(1,2)
(841.80,3)
};
\addplot+[xbar,blue] coordinates { % min local CNN
(82.08, 0)
(124.05, 1)
(197.45, 2)
(1, 3)
};
\addplot+[xbar,darkblue] coordinates { % max local CNN -- just difference max-min
(8.79, 0)
(12.15, 1)
(7.43, 2)
(1, 3)
};
\addplot+[xbar,cyan] coordinates { % global coarse CNN
(4.14, 0)
(4.15, 1)
(4.26, 2)
(1, 3)
};
\legend{Global CNN, Min Local CNN, Max Local CNN, Global Coarse DNN}
\end{axis}
\end{tikzpicture}
\caption{Comparison of the training times for the global CNN models and our proposed decomposed approach for the TF-Flowers dataset~\cite{tfflowers} for the ResNet20. We show in teal the training time required for the global 2D CNN model using the entire TF-Flowers images as input data. See~\cref{fig:runtime_flowers} for color explanations. 
See~\cref{tab:res_acc_flowers_ResNet20} for the corresponding classification accuracies of the trained models. }	
\label{fig:runtime_flowers_ResNet20}
\end{figure}

\subsection{Experiments on Extended Yale Face Database B}
\label{sec:re-faces}

Next, we test our approach for a more practically relevant image classification problem in form of face recognition. As an exemplary dataset for this application, we choose the Extended Yale Face Database B~\cite{Yale_faces1,Yale_faces2}. This database consists of gray-scale images of $38$ different individuals with a size of $192\times168$ pixels; see also~\cref{sec:data_nets_faces}.
In general, the task of face recognition is relatively hard and requires the availability of very large quantities of training data and/or the training of highly complex network architectures to obtain satisfying generalization properties~\cite{parkhi:2015:deepface}.
In our experiments, we have experienced a significant variance in the related accuracy values of both the global CNN and the local networks for different initializations of the network weights as well as for different splits of the dataset into training and validation data. 
However, given that the goal of this work is not to train a (new) global machine learning model with the highest possible accuracy value but to compare the performance of our proposed two-level CNN-DNN architecture with the training of a given global CNN model, the obtained results still deliver a good impression of how well the proposed training approach performs in comparison to the global training.

In~\cref{tab:res_acc_faces}, we provide the classification performances for the global and the decomposed training strategy for a fixed splitting of the data into training and validation data using a fixed random seed and a given initialization of the network weights as uniformly distributed values in $[0,1]$.
For this case, the global CNN shows a very high accuracy in the classification of both, the training and the validation data. Even though the validation accuracy values for the local subnetworks are, on average, remarkably lower than for the global CNN, the final validation accuracy of the CNN-DNN model is in a similar order of magnitude as for the global CNN and only slightly lower. 

\begin{table}[th]
\centering
\scalebox{0.94}{
\begin{tabular}{l||c||c|c|c||c}
  {\bf Decomp.} & {\bf global CNN} & \multicolumn{3}{c||}{\bf $p^2$ local CNN} & {\bf CNN-DNN} \\\hline 
  & & avg & min & max & \\\hline
type A & 0.9931 & 0.6353 & 0.0178 & 0.9360 & 0.9677 \\
 $3\times 3$, $\delta=0$ & (0.9935) & (0.7495) & (0.0267) & (0.9953) & (1.0000) \\\hline
type A  & 0.9931 & 0.5601 & 0.0255 & 0.9336  & 0.9763 \\
 $4\times 4$, $\delta=0$  & (0.9935) & (0.6981) & (0.0267) & (0.9905) & (1.0000) \\\hline
\end{tabular}
}
\caption{\label{tab:res_acc_faces} Classification accuracy values for the validation and the training data (in brackets) for the global CNN and the proposed parallel training approach, i.e., the local CNNs and the final classification obtained by the CNN-DNN approach for the Extended Yale Face Database B~\cite{Yale_faces1,Yale_faces2}. See~\cref{sec:data_nets_faces} for a detailed description of the dataset and the trained 2D CNN architectures.}
\end{table}

\subsection{Experiments on Chest CT Scans}
\label{sec:res-CT}

Finally, we test our approach for a 3D CNN model which is trained for a binary classification problem of chest CT data; see~\cref{sec:data_nets_CT}. 
The 3D CNN model uses as input volumetric data of $128\times128\times64$ voxels.

First, in~\cref{tab:res_acc_CT}, we present the classification accuracies for the global 3D CNN, the local 3D subnetworks and the CNN-DNN using the additional dense feedforward global coarse net. In analogy to the previous experiments for 2D CNNs, the performance of the local subnetworks decreases in comparison to the global CNN and with an increasing number of subnetworks. However, comparing the performance of the decomposed CNN-DNN approach to the global CNN, the decomposed CNN-DNN approach achieves a higher classification accuracy both on the training and the validation data for all tested decompositions.

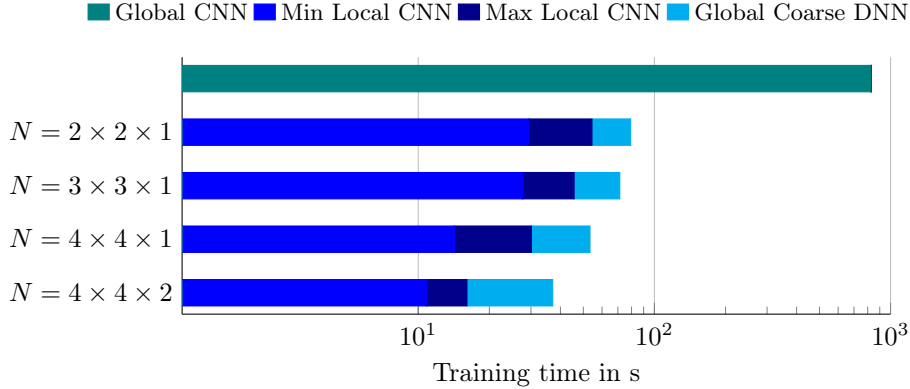
\begin{figure}
\centering
\begin{tikzpicture}
\begin{axis} [xbar stacked,width=11cm,height=5cm, axis y line*=none,
    axis x line*=bottom, ytick={0,1,2,3}, yticklabels={$N=4\times4\times2$, $N=4\times 4 \times 1$, $N=3\times 3 \times 1$, $N=2\times 2 \times 1$, Global}, xlabel = Training time in s, xmajorgrids=true, xmode= log, xmin=1, xmax=1000, xtick={10,100,1000}, enlarge x limits=0.0, legend style={legend columns=4,at={(0.45,1.25)},anchor=north,draw=none, font=\small}]
\addplot+[xbar,teal] coordinates { % global CNN
(1,0)
(1,1)
(1,2)
(1,3)
(825.85,4)
};
\addplot+[xbar,blue] coordinates { % min local CNNs
(9.86, 0)
(13.24, 1)
(26.67, 2)
(28.27, 3)
(1, 4)
};
\addplot+[xbar,darkblue] coordinates { % max local CNNs - just difference max-min
(5.22, 0)
(15.92, 1)
(18.04, 2)
(25.14, 3)
(1, 4)
};
\addplot+[xbar,cyan] coordinates { % global coarse CNN 
(21.03, 0)
(23.21, 1)
(25.68, 2)
(24.95, 3)
(1, 4)
};
\legend{Global CNN, Min Local CNN, Max Local CNN, Global Coarse DNN}
\end{axis}
\end{tikzpicture}
\caption{Comparison of the training times for the global CNN models and our proposed decomposed approach for the chest CT dataset~\cite{Chest_CT}. We show in teal the training time required for the global 3D CNN model using the entire CT data as input data. See~\cref{fig:runtime_flowers} for color explanations. 
See~\cref{tab:res_acc_CT} for the corresponding classification accuracies of the trained models. }	
\label{fig:runtime_CT}
\end{figure}

Second, in~\cref{fig:runtime_CT}, we provide the (maximum) training times for the global CNN, the local subnetworks, and the CNN-DNN approach for different decompositions of the data into subsets of voxel data, that is, different numbers of subnetworks using the same computing resources. As before, we always train the global network on one GPU and equally distribute the training of the local CNNs to the $8$ GPUs. More precisely, the training of the $k$-th local CNN is assigned to GPU with index=$k \mod 8$. Subsequently, we train the global coarse DNN on one GPU. All networks are trained for $500$ epochs and using an early stopping criterion~\cite{prechelt:1998:early_stop} with respect to the validation accuracy with a patience of $15$ epochs.

Similar to the 2D CNN experiments in~\cref{fig:runtime_flowers}, the maximum training time for the local subnetworks decreases with an increasing number of subnetworks while the training time for the global coarse net remains approximately constant for all four tested decompositions.
In sum, the total training time of the global 3D CNN model can be reduced by a factor of $10.64$ compared to the decomposition into $2\times 2 \times 1$ local 3D subnetworks. This is an even more drastical reduction of the required training time than for the 2D CNN model in~\cref{sec:res-flowers}. For the decomposition into $4\times 4\times 1$ subnetworks, the training time for the global CNN is $15.93$ larger than for the decomposed strategy and by a factor of $22.87$ larger for a decomposition into $4\times  4 \times 2$ subnetworks.

\begin{table}[th]
\centering
\scalebox{0.94}{
\begin{tabular}{l||c||c|c|c||c}
  {\bf Decomp.} & {\bf global CNN} & \multicolumn{3}{c||}{\bf $p^2$ local CNN} & {\bf CNN-DNN} \\\hline 
  & & avg & min & max & \\\hline
type A & 0.7667 & 0.6875 & 0.6667 & 0.7000 & 0.9143 \\
$2\times 2 \times 1$, $\delta=0$ & (0.8214) & (0.7780) & (0.6929) & (0.9214) & (0.9357)  \\\hline
type A & 0.7667 & 0.6492 & 0.5833 & 0.7333 & 0.8986 \\
$3\times 3\times 1$, $\delta=0$ & (0.8214) & (0.6651) & (0.5929) & (0.7286) & (0.9514) \\\hline
type A & 0.7667 & 0.6366 & 0.5333 & 0.7667 & 0.8571 \\ 
$4\times 4\times 1,$ $\delta=0$ & (0.8214) & (0.6409) & (0.5071) & (0.8571) & (0.9357) \\\hline
type A & 0.7667 & 0.6479 & 0.5585 & 0.7847 & 0.8633 \\ 
$4\times 4\times 1,$ $\delta=4$ & (0.8214) & (0.6409) & (0.5322) & (0.8667) & (0.9456) \\\hline
type A & 0.7667 & 0.6349 & 0.5702 & 0.7480 & 0.8988 \\
$4\times4\times2$, $\delta=0$ & (0.8214) & (0.6424) & (0.5837) & (0.7675) & (0.9493) \\\hline
\end{tabular}
}
\caption{\label{tab:res_acc_CT} Classification accuracy values for the validation and the training data (in brackets) for the global CNN and the proposed parallel training approach, i.e., the local CNNs and the final classification obtained by the CNN-DNN approach for the Chest CT scan dataset~\cite{Chest_CT}. See~\cref{sec:data_nets_CT} for a detailed description of the dataset and the trained 3D CNN architectures.}
\end{table}

\section{Conclusion}
\label{sec:concl}

In the present paper, we have introduced a novel CNN-DNN architecture which is loosely inspired by two-level domain decomposition methods and which supports a model parallel training strategy. 
We have applied the proposed architecture to different image recognition problems. The provided results show that the proposed approach can significantly accelerate the required training time compared to the training of a global CNN model. Additionally, the experimental results indicate that the method can also often help to improve the accuracy of the classification.

\section*{Acknowledgements} We gratefully acknowledge the use of the computational
facilities of the Center for Data and Simulation Science (CDS) at the University of
Cologne. 
Additionally, we gratefully acknowledge the use of the CIFAR-10 dataset \cite{Cifar10_TR}, the TF-Flowers dataset~\cite{tfflowers}, the Extended Yale Face Database B~\cite{Yale_faces1,Yale_faces2}, and the MosMedData~\cite{Chest_CT} to evaluate our proposed machine learning approach.

\bibliographystyle{siamplain}
\bibliography{dnn_cnn} 

\end{document}